\documentclass{article}

\usepackage{microtype}
\usepackage{graphicx}
\usepackage{subcaption}
\usepackage{booktabs} 
\usepackage{float}

\usepackage{multirow}
\usepackage{colortbl} 
\usepackage{xcolor}
\definecolor{graybg}{rgb}{0.9, 0.9, 0.9}

\usepackage{hyperref}

\usepackage[preprint]{icml2026}

\usepackage{amsmath}
\usepackage{amssymb}
\usepackage{mathtools}
\usepackage{amsthm}

\usepackage[capitalize,noabbrev]{cleveref}
\usepackage{enumitem}

\theoremstyle{plain}
\newtheorem{theorem}{Theorem}[section]
\newtheorem{proposition}[theorem]{Proposition}
\newtheorem{lemma}[theorem]{Lemma}

\theoremstyle{definition}
\newtheorem{definition}[theorem]{Definition}

\theoremstyle{remark}

\definecolor{RED}{rgb}{0.7,0,0}
\definecolor{BLUE}{rgb}{0,0,0.69}
\definecolor{GREEN}{rgb}{0,0.6,0}
\definecolor{PURPLE}{rgb}{0.69,0,0.8}
\definecolor{BLACK}{rgb}{0,0,0}
\definecolor{TEAL}{rgb}{0, 0.5, 0.5}

\newcommand{\BLUE}{\color[rgb]{0,0,0.69}}

\DeclareMathOperator{\rref}{ref}
\DeclareMathOperator{\gen}{gen}

\usepackage[normalem]{ulem}

\usepackage[textsize=tiny]{todonotes}

\icmltitlerunning{Latent Iterative Refinement Flow: A Geometric Constrained Approach for Limited-Data Generation}

\begin{document}

\twocolumn[
  \icmltitle{
  \textcolor{BLUE}{L}atent
  \textcolor{BLUE}{I}terative
  \textcolor{BLUE}{R}efinement
  \textcolor{BLUE}{F}low:
  A Geometric Constrained Approach \\for Limited-Data Generation}



  \icmlsetsymbol{equal}{*}

\begin{icmlauthorlist}
    \icmlauthor{Songtao Li}{hust_math,hust_cms}
    \icmlauthor{Tianqi Hou}{huawei}
    \icmlauthor{Zhenyu Liao}{hust_eic}
    \icmlauthor{Ting Gao}{hust_math,hust_cms}
\end{icmlauthorlist}

\icmlaffiliation{hust_math}{School of Mathematics and Statistics, Huazhong University of Science and Technology, Wuhan 430074, China}
\icmlaffiliation{hust_eic}{School of Electronic Information and Communications, Huazhong University of Science and Technology, Wuhan, China}
\icmlaffiliation{hust_cms}{Center for Mathematical Science, Huazhong University of Science and Technology, Wuhan 430074, China}
\icmlaffiliation{huawei}{Huawei} %

  \icmlcorrespondingauthor{Zhenyu Liao}{zhenyu\_liao@hust.edu.cn}
  \icmlcorrespondingauthor{Ting Gao}{tgao0716@hust.edu.cn}

  \icmlkeywords{Machine Learning, ICML}

  \vskip 0.3in
]



\printAffiliationsAndNotice{}  

\begin{abstract}
Diffusion and flow-matching models trained with limited data often tend to memorize the training data instead of generalization, leading to severely reduced diversity. 
In this paper, we provide a dynamical perspective and identify this ``collapse-to-memorization'' phenomenon as a consequence of the \emph{velocity field collapse}, where the learned field degenerates into isolated point attractors and trap the sampling trajectories.
Inspired by this novel view, we introduce \textbf{{\BLUE L}atent {\BLUE I}terative {\BLUE R}efinement {\BLUE F}low ({\BLUE LIRF})}, a geometry-aware framework for from-scratch training of diffusion models in the limited-data regime. 

By exploiting the intrinsic geometry of a semantically aligned latent space, LIRF progressively densifies the training data manifold via a \emph{generation--correction--augmentation} closed loop, thereby effectively resolving the velocity field collapse.
Theoretical guarantee on the convergence of this manifold densification procedure is also provided.
Experiments on FFHQ subsets and Low-Shot datasets demonstrate the advantageous performance of LIRF over existing diffusion models for limited-data generation, achieving significantly higher diversity and recall, with comparably good generative performance.
\end{abstract}

\section{Introduction}

Diffusion and flow-matching 
models have emerged as the dominant paradigms for high-fidelity image generation, owing to their superior training stability and favorable convergence properties~\citep{ho2020denoising,song2020score,karras2022elucidating,Rombach_2022_CVPR,lipman2022flow}. 
Despite algorithmic differences, recent studies have established a unifying perspective: both diffusion and flow-matching models can be viewed as learning a \emph{time-dependent velocity field} that defines a probability path transporting samples from noise to data~\citep{lipman2024flow,ma2024sit,li2025back}. 
This formulation provides useful explanations for the stability and scalability that enable diffusion-based models to achieve superior performance across a wide range of generative tasks~\cite{Rombach_2022_CVPR,song2023consistency,podell2023sdxl,peng2025open}.

Despite these successes, diffusion models typically rely on \emph{large and diverse training datasets}. 
In the limited-data regime, they are often observed to struggle to learn a globally consistent and generalizable velocity field, and may instead reproduce training samples with high fidelity—a phenomenon known as \emph{memorization}~\citep{yoon2023diffusion,zhang2023emergence,kadkhodaie2023generalization,favero2025bigger}. 
However, assembling large-scale, high-quality, and diverse datasets is frequently difficult in practice, particularly in domain-specific or resource-constrained settings~\citep{karras2020training,zhao2020differentiable}. 
Improving the data efficiency of diffusion-based generative models therefore remains a fundamental and practically important challenge.

Existing methods for diffusion training under data-scarcity typically fall into two main categories: 
\emph{fine-tuning pretrained models} and \emph{data augmentation}. 
The fine-tuning approach adapts diffusion models pretrained on large-scale datasets to a target domain using only a small number of target-domain samples~\citep{ruiz2023dreambooth,zhang2023adding,moon2022fine,yang2024fewshot}. 
Its effectiveness, however, depends heavily on both the availability of strong pretrained models and the degree of semantic overlap between the source and target domains, which limits its applicability when training from scratch is required.  

Another line of work focuses on data augmentation. 
Adaptive augmentation strategies such as ADA~\citep{karras2020training} and differential augmentation~\citep{zhao2020differentiable} aim to mitigate overfitting under limited data 
by introducing stochastic pixel-space transformations, including geometric transformations, color jittering, and noise injection. 
While these methods can improve generalization to some extent, they primarily operate in pixel space and, as we shall see in \Cref{fig:spiral_compaire}, often fail to leverage the intrinsic geometric structure of the latent space.

Unlike prior approaches, we address limited-data diffusion training from a geometric perspective.
We first embed the limited training samples into a semantically structured latent space, and then progressively densify the resulting latent manifold via an iterative refinement flow. 
This yields \textbf{{\BLUE L}atent {\BLUE I}terative {\BLUE R}efinement {\BLUE F}low ({\BLUE LIRF})}, a theoretically grounded framework for diffusion training that improves data efficiency by learning generative dynamics over a refined latent distribution.

LIRF is built upon two key principles.
First, manifold densification must be carried out in a semantically meaningful latent space; however, conventional VAEs~\citep{pinheiro2021variational,Rombach_2022_CVPR} typically prioritize compression and pixel-level reconstruction without an explicit latent structure.
To address this limitation, we embed training data using DiNO-VAE~\cite{xu2025exploring}, which yields a semantically aligned latent representation in which distances correspond to meaningful semantic dissimilarities. 
This ensures that manifold densification is performed in a space that preserves semantic structure.

Second, in this semantic latent space, LIRF implements a \emph{generation--correction--augmentation} closed loop that enables progressive refinement of the latent generative process. 
In the early stages of training, when only limited data are available, the model may generate samples that drift away from the latent data manifold. 
LIRF refines such samples through a geometric \emph{correction operator}, which contracts them towards the local data manifold, thereby maintaining manifold consistency under data scarcity.
By iteratively generating, correcting, and augmenting training data, LIRF progressively densifies the latent manifold, which increasingly approximates the true data manifold.


\subsection{Our contributions}
Our main contributions can be summarized as follows.
\begin{enumerate}
    \item We characterize velocity field collapse as an important failure mechanism under data scarcity. We demonstrate that training samples act as 
    local attractors that trap nearby sampling trajectories, providing a dynamical perspective on why memorization occurs.
    
    \item We propose LIRF, a geometry-aware flow-matching training framework that progressively densifies sparse data manifolds via a \emph{generation–correction–augmentation} loop, where a locally contractive correction operator pulls sampled candidates back toward the data manifold.
    
    \item We provide both theoretical convergence guarantees and experiments on FFHQ and Low-Shot \cite{zhao2020differentiable} datasets demonstrating that LIRF substantially improves performance of diffusion baselines.
\end{enumerate}

\subsection{Related Work}

\paragraph{A unified view of diffusion and flow-matching models.}
Recent work has established a unified perspective on diffusion and flow-based generative models by viewing both as learning a time-dependent velocity field that transports probability mass from noise to data~\cite{albergo2025stochastic,ma2024sit,li2025back}.
Although diffusion models were originally formulated via stochastic differential equations, their dynamics admit an equivalent deterministic probability flow ODE formulation~\cite{ho2020denoising,song2020score}. 
Building upon this equivalence, flow matching proposes a simulation-free objective for directly regressing velocity fields along prescribed probability paths, with standard diffusion as a special case~\cite{lipman2022flow}.   
Recent studies further demonstrate that optimizing the geometry of transport paths (such as through rectified or straightened trajectories) can substantially improve sampling efficiency~\cite{liu2022flow,wang2024rectified,ma2024sit}. 
In contrast to prior work that focuses on global path geometry, our method leverages latent-space geometry to enable data-efficient refinement under limited training data.

\paragraph{Memorization in diffusion model.}
Diffusion models are prone to memorizing training data, particularly in over-parameterized or data-scarce regimes~\cite{carlini2023extracting,somepalli2023diffusion,gu2023memorization}. 
This behavior is commonly interpreted as a form of overfitting driven by excessive model capacity~\cite{yoon2023diffusion,kadkhodaie2023generalization}. 
While under-parameterized models tend to learn dataset-agnostic score functions, sufficiently expressive models can encode dataset-specific details, leading to \emph{training sample reproduction}~\cite{kadkhodaie2023generalization}. 
Recent geometric analyses further suggest that limited data induces manifold collapse and a shrinking generalization window during training~\cite{achilli2024losing}. 
These findings motivate our use of geometric correction and iterative refinement to stabilize generative dynamics under data scarcity.

\paragraph{Diffusion models with limited data.} 
Despite their success at scale, diffusion models typically rely on large and diverse datasets. 
Most existing approaches address data scarcity via transfer learning, by fine-tuning models pretrained on similar domains~\cite{ruiz2023dreambooth,yang2024fewshot}. 
In contrast, training diffusion models from scratch under limited data remains relatively underexplored. 
Representative efforts include Patch Diffusion~\cite{wang2023patch}, which reduces data requirements through patch-wise training, and LD-Diffusion~\cite{zhang2025training}, which restricts the denoiser's hypothesis space and employs mixed augmentations. 
However, these methods largely overlook the \emph{intrinsic geometric structure} of the data manifold; our work addresses this gap by explicitly exploiting the semantic latent manifold geometry to progressively densify the data distribution through iterative refinement.


\section{Preliminaries}
\label{sec:preliminaries}

While the proposed framework is applicable to a broad class of diffusion models, we instantiate it using a latent flow-matching model as the generative backbone ~\cite{Rombach_2022_CVPR,esser2024scaling,labs2025flux}.
Let $z \in \mathbb{R}^d$ denote the latent representation of a data sample $x$, obtained from a frozen pretrained encoder $z=E(x)$. 
Flow matching formulates generative modeling as learning a time-dependent vector field $v_\theta(z_t,t)$ that transports samples from a prior distribution $q_{0}$ (e.g., $\mathcal{N}(0,I)$) to the data distribution $q_1$.
The generation process is governed by the ODE:
\begin{equation}
\frac{d z_t}{d t} = v_\theta(z_t, t),
\label{eq:fm_latent_ode}
\end{equation}
with initial condition $z_0 \sim \mathcal{N}(0,I)$.
The velocity field $v_\theta$ is trained by minimizing the conditional flow matching objective~\cite{tong2023improving}:
\begin{equation*}
\mathcal{L}_{\text{CFM}}(\theta)
= \mathbb{E}_{t \sim \mathcal{U}[0,1],\, z \sim q_t(\cdot \mid z_1)}
\left\| v_\theta(z,t) - u_t(z \mid z_1) \right\|_2^2,
\end{equation*}
where $u_t(z \mid z_1)$ denotes the conditional velocity field associated with the probability path $q_t(z \mid z_1)$.

Following common practice~\cite{liu2022flow,tong2023improving}, we adopt here a linear probability path
\begin{equation}
z_t = t z_1 + (1 - t) z_0
\label{eq:straight_path}
\end{equation}
which yields a constant target velocity $u_t(z \mid z_1) = z_1 - z_0$.
The effectiveness of flow matching critically relies on the assumption that the target distribution $q_1$ is sufficiently well sampled. 
In limited-data regimes, this assumption no longer holds, causing the learned transport dynamics to collapse.
We identify this phenomenon as a primary bottleneck for flow-based generative models under data scarcity, and analyze it in the following section.

We use flow matching as the generative backbone for two practical reasons.
First, flow matching supports deterministic ODE sampling with straight transport paths, often requiring fewer numbers of function evaluations (NFEs) than standard diffusion sampling and reducing the computation of interleaved candidate harvesting \cite{dao2023flow,lipman2022flow,ma2024sit}.
Second, $v$-prediction can provide a more stable target than $\epsilon$-prediction in diffusion models, which can improve training efficiency in limited-data \cite{li2025back}.

\subsection{Velocity Field Collapse under Data Scarcity}
\label{sub:degeneration_of_velocity_fields_under_data_scarcity}

While performance degradation under limited data is often attributed to memorization or overfitting~\cite{gu2023memorization,kadkhodaie2023generalization}, here we adopt a dynamical perspective:
the learned time-dependent velocity field degenerates into a collection of isolated point attractors.
We refer to this failure mode as \emph{velocity field collapse}, in which the sampling distribution concentrates around
a small number of point masses around training samples, i.e., $q_1^\theta(z) \approx \sum_{i=1}^N \pi_i\,\delta(z-z_i)$, thereby severely limiting sample diversity.

\begin{figure}[t]
    \centering
    \includegraphics[width=0.95\columnwidth]{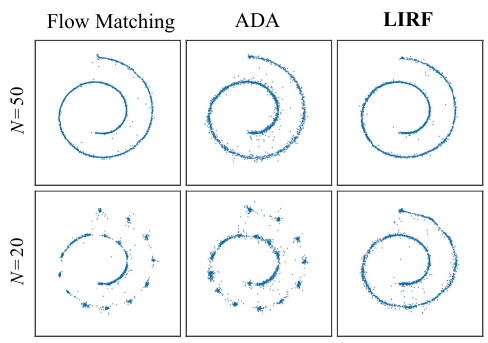}
    \caption{{\textbf{Illustration of velocity field collapse for vanilla flow matching, ADA, and the proposed LIRF on 2D spiral data.}
    As the number of training samples decreases, vanilla flow matching degrades towards ``memorization'' of isolated training samples, while ADA produces fragmented generations with disconnected regions. 
    In contrast, LIRF preserves a continuous generative distribution that remains aligned with the underlying spiral manifold.}}
    \label{fig:spiral_compaire}
    \vspace{-0.1in}
\end{figure}

\begin{figure}[b]
\centering
\includegraphics[width=0.99\columnwidth]{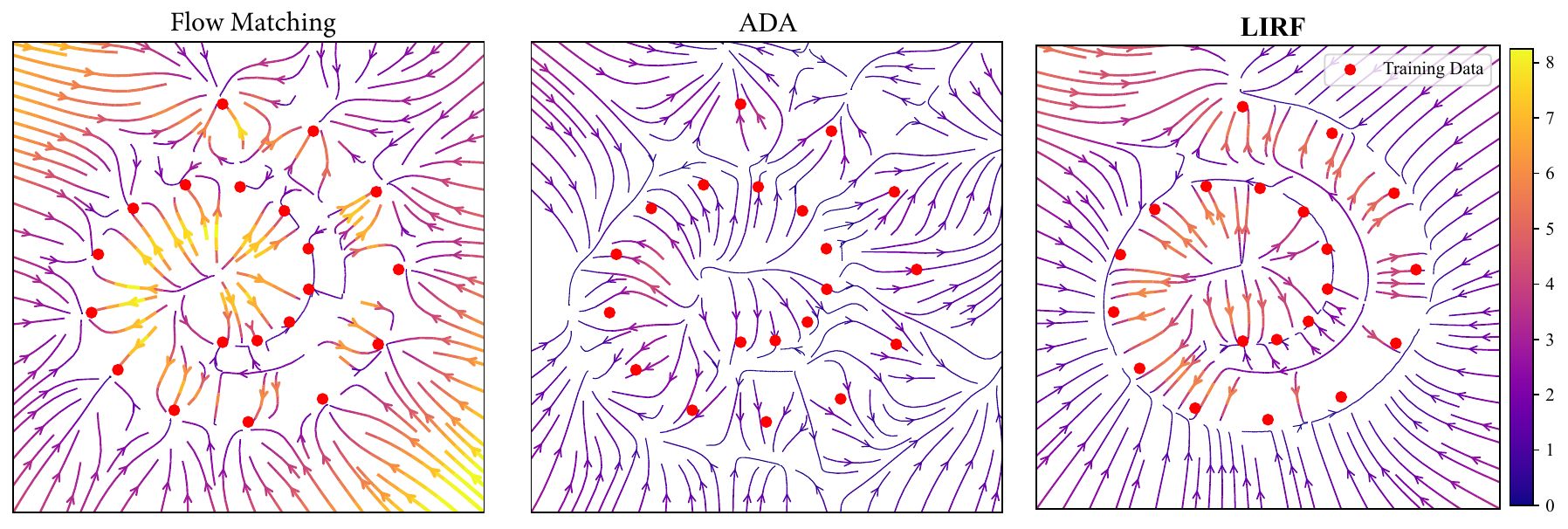} 
    \caption{{\textbf{Visualization of learned velocity fields in the setting of \Cref{fig:spiral_compaire}.} 
    Arrows indicate the direction and magnitude of the learned velocity field, while \textbf{\textcolor{red}{red}} dots denote training samples.}}
    \label{fig:velocity_compare}
    \vspace{-0.1in}
\end{figure}

\begin{figure*}[t]
\centering
\includegraphics[width=0.99\textwidth]{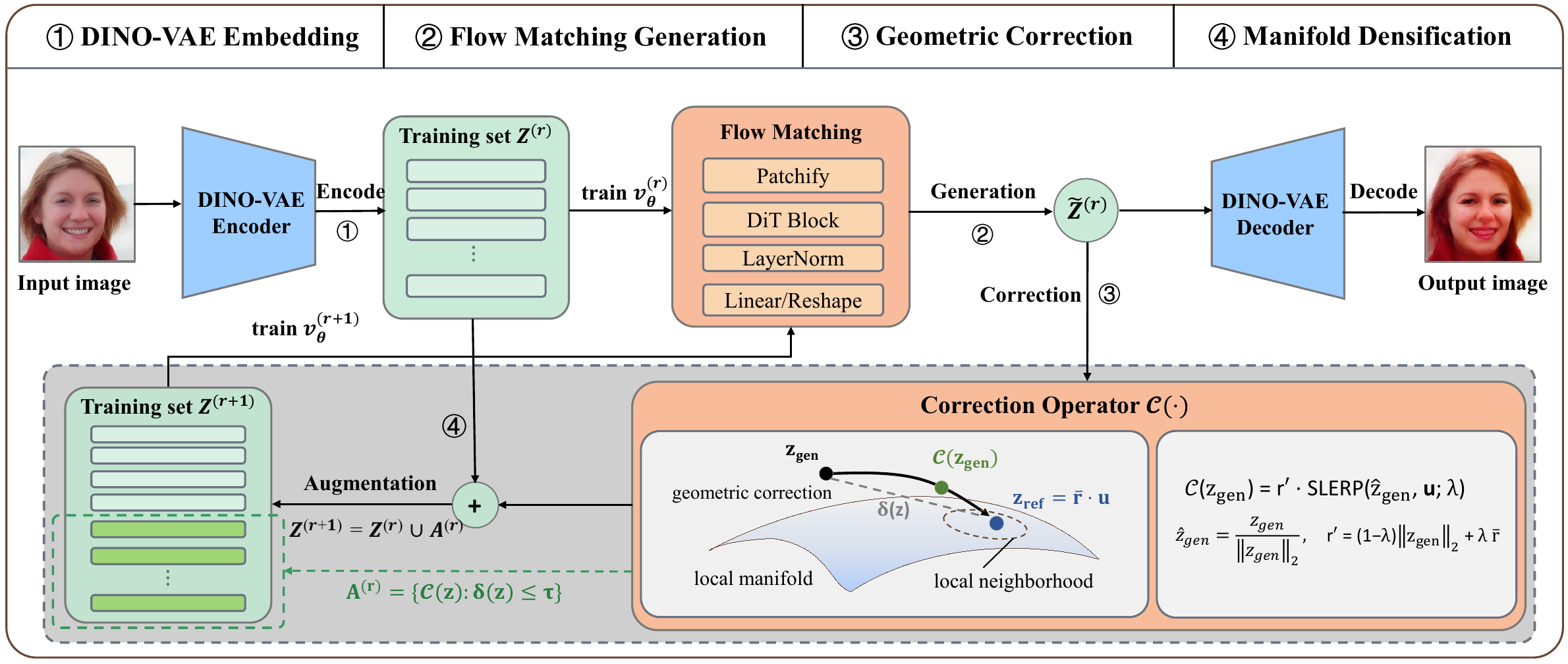}
    \caption{{\textbf{Overview of Latent Iterative Refinement Flow (LIRF).} 
    (1) Training images are encoded by a frozen DiNO-VAE into a semantically aligned latent space, forming the current training set $Z^{(r)}$. 
    (2) A flow-matching model is trained on $Z^{(r)}$ and periodically generates a candidate set $\tilde Z^{(r)}$ via ODE sampling. 
    (3) Each candidate $\tilde{z}$ is refined using a geometric correction operator $\mathcal{C}(\cdot)$, which pulls it towards a local reference point $z_{\rref}$. 
    (4) Corrected samples with small correction magnitude ($\delta(z)=|\tilde z-\mathcal{C}(\tilde z)|_2\leq\tau$) are admitted into $\mathcal{A}^{(r)}$ and merged into the training set to form $Z^{(r+1)}$.
    This generation--correction--augmentation loop progressively densifies the latent manifold and mitigates velocity field collapse under data scarcity.}}
    \label{fig:LIRF_framework}
    \vspace{-0.1in}
\end{figure*}

To visualize this phenomenon, we compare in \Cref{fig:spiral_compaire} samples generated by vanilla flow matching, ADA, and the proposed LIRF, when trained on a 2D spiral dataset with limited sample size ($N  = 20$ and $50$). 
As the amount of training data decreases, vanilla flow matching exhibits clear signs of collapse: generated samples concentrate tightly around individual training samples rather than forming a coherent continuous structure.
While ADA introduces stochastic data augmentation through local transformations, it is \emph{insufficient} to bridge these isolated regions, resulting in a fragmented latent manifold that lacks global coherence.

We further visualize the corresponding velocity fields in \Cref{fig:velocity_compare}. 
Instead of forming a smooth manifold-aligned velocity field, the fields learned by vanilla flow matching and ADA exhibit sharp peaks at or around training samples. 
These peaks act as \emph{point attractors}: once a generative trajectory enters their basins of attraction, it is drawn towards a specific training example, while regions between samples display weak or incoherent velocities that prohibit transport across low-density gaps. To further verify that this phenomenon persists beyond toy data, we provide additional evidence on FFHQ in \Cref{app:ffhq_velocity_collapse} of the appendix.

These dynamics indicate that, under data scarcity, the learned velocity field often exhibit structural fragmentation, characterized by disconnected attractor regions that locally trap generative trajectories and bias generated samples toward training instances. 
Such behavior \emph{cannot} be effectively addressed through regularization or pixel-space data augmentation \emph{alone}. 
While these techniques may expand local neighborhoods around individual samples, they do \emph{not} modify the \emph{global} structure of the learned field or restore connectivity across low-density regions. 
This observation motivates explicitly leveraging manifold geometry to stabilize generative dynamics in limited-data regimes.


\section{Method: Latent Iterative Refinement Flow}

In this section, we present the proposed LIRF framework for diffusion training under limited data. 
Specifically, LIRF first embeds the limited training samples into a semantically aligned latent space, and then performs iterative latent manifold densification through a generation--correction--augmentation closed loop. 
These two components are described in \Cref{sub:semantically_aligned_latent_space} and \Cref{sub:iterative_latent_manifold_densification}, respectively; see \Cref{fig:LIRF_framework} for an overview.

\subsection{Embedding in Semantically Aligned Latent Space}
\label{sub:semantically_aligned_latent_space}


LIRF aims to mitigate velocity field collapse by progressively densifying the initial training support within a predefined latent space. 
This approach relies on the following geometric properties of the chosen latent space:
\begin{enumerate}[leftmargin=*]
    \item \emph{Neighborhood preservation}: semantically similar images in the training set $X$ remain close in the latent space $Z$ (e.g., under cosine similarity);  
    \item \emph{Meaningful interpolation}: locally, the regions between nearby samples should admit semantically valid interpolations when decoded back to the image space $X$. 
\end{enumerate}

\begin{figure}[t]
\centering
\includegraphics[width=0.99\columnwidth]{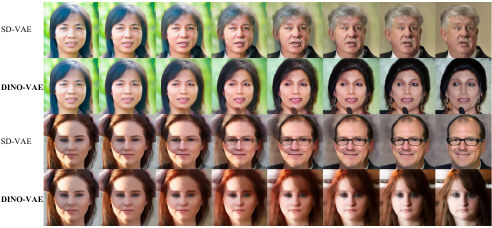}
    \caption{\textbf{Top/Third Rows (SD-VAE):} Interpolation trajectories traverses semantically implausible region, resulting in severe \emph{ghosting artifacts}, blurriness, and incoherent facial structures.
    \textbf{Second/Bottom Rows (DiNO-VAE):} Intermediate samples remain sharp and structurally coherent, confirming that the latent space exhibits approximate semantic convexity under local interpolation.}
    \label{fig:vae_comparison}
    \vspace{-0.1in}
\end{figure}

However, standard VAEs (e.g., SD-VAE~\cite{Rombach_2022_CVPR}) are primarily optimized for image compression and pixel-level reconstruction, rather than for preserving semantic geometry in the latent space.
As a result, the induced latent space often lacks local semantic consistency. 
In particular, linear interpolations between latent representations $z_i$ and $z_j$ may traverse semantically implausible regions and decode into fragmented or incoherent artifacts. 
As illustrated in \Cref{fig:vae_comparison}, this manifests as \emph{ghosting} effects and semantic discontinuities, indicating that densification in such a latent space would primarily inject noise rather than meaningfully expand meaningful data coverage.

To overcome this limitation, we operate in the latent space induced by DiNO-VAE~\cite{xu2025exploring}. 
Prior work has shown that representations learned by DINOv2~\cite{oquab2023dinov2} exhibit an approximately hyperspherical geometry, in which semantic concepts form coherent clusters and cosine distance correlates well with semantic similarity~\cite{oquab2023dinov2,govindarajan2024dino}. 
This structured geometry enables the use of cosine-based neighborhoods and spherical interpolation (in $\mathcal{Z}$) that remain ``aligned'' with the underlying data manifold (in $\mathcal{X}$), thereby supporting subsequent stable and semantically meaningful latent densification. 
The DiNO-VAE rows in \Cref{fig:vae_comparison} empirically confirm that unlike standard VAEs, interpolations in the DiNO-aligned latent space preserve perceptual continuity.




\subsection{Iterative Latent Manifold Densification}
\label{sub:iterative_latent_manifold_densification}

Within this semantically aligned latent space, the limited-data generation problem can be viewed from a geometric perspective.
Starting from a sparse anchor set $Z^{(0)}= \{ z_i \}_{i=1}^N$, our goal is to progressively densify its effective support so as to facilitate learning a global and coherent transport field. 
When training directly on such a small anchor set, the resulting signal likely is both weak and highly localized, which often leads to degraded sample quality and the collapse behaviors discussed in \Cref{sub:degeneration_of_velocity_fields_under_data_scarcity}. 
We therefore introduce an \emph{iterative refinement strategy} that gradually expands the effective training support while preserving fidelity to the underlying data manifold.

LIRF proceeds through three steps: \textbf{generation}, \textbf{correction}, and \textbf{augmentation}.
At refinement round $r$, we maintain a training set $Z^{(r)}$ and train a flow matching model with velocity field $v_{\theta}^{(r)}$ on $Z^{(r)}$. 
We then sample candidate points from $v_{\theta^{(r)}}$, refine them via the correction operator, and augment $Z^{(r)}$ with admitted corrected samples to get the the updated set $Z^{(r+1)}$.
This closed-loop procedure progressively densifies the latent manifold and aims to stabilize generative training under data scarcity.

Below, we describe each of the three steps in detail.

\noindent\textbf{Step 1: Generation.} 
Unlike conventional pipelines that sample only \emph{after} training converges, LIRF generates samples \emph{during} training.
Specifically, every $\Delta$ steps, we trigger a refinement round and draw candidate samples using a snapshot of the current model.
Formally, at round $r$, candidates points are obtained by integrating the probability-flow ODE associated with velocity field $v_{\theta}^{(r)}$:
\begin{equation}
\tilde{z} = z_0 + \int_0^1 v_{\theta}^{(r)}(z_t, t) \, dt, \quad z_0 \sim \mathcal{N}(0, I),
\label{eq:generation_ode}
\end{equation}
yielding a candidate set
$\tilde{Z}^{(r)}=\{\tilde{z}_k^{(r)}\}_{k=1}^{M}$ at round $r$.

Importantly, generation at this stage is \emph{not} intended to produce high-fidelity samples. 
Instead, its purpose is to explore semantically plausible latent regions that remain under-covered by the current anchor set $Z^{(r)}$.
Prior work suggests that early-stage generative models often capture coarse global structure and yield diverse layouts, despite lacking fine-grained details~\cite{gu2023memorization,favero2025bigger}.
By periodically generating candidate samples, LIRF leverages this exploratory behavior to identify and bridge semantic gaps between existing anchors.

\paragraph{Step 2: Correction.} 
The raw candidate set $\tilde{Z}^{(r)}$ generated in Step~1 via Eq.~\eqref{eq:generation_ode} is exploratory in nature and may deviate from the data manifold, particularly during the early stages of training. 
Admitting such candidates directly would introduce noise and geometric inconsistency into the training set. 
We therefore introduce a geometric \emph{correction operator} $\mathcal{C}(\cdot)$ that pulls generated samples towards the local neighborhood of existing latent anchors.

\begin{table*}[t]
    \centering
    \small
    \caption{Comparison of generation quality and diversity on FFHQ subsets ($256\times256$) using $50$K generated samples. We report FID ($\downarrow$), Precision ($\uparrow$), and Recall ($\uparrow$). The best results are shown in \textbf{bold}, and the second-best results are \underline{underlined}. }
    \label{tab:ffhq_results}
    \setlength{\tabcolsep}{3.5pt}
    \renewcommand{\arraystretch}{1.1}

    \begin{tabular}{@{}l p{5.6cm} ccc ccc ccc@{}}
        \toprule
        & & \multicolumn{3}{c}{\textbf{FFHQ-100}} & \multicolumn{3}{c}{\textbf{FFHQ-1k}} & \multicolumn{3}{c}{\textbf{FFHQ-2k}} \\
        \cmidrule(lr){3-5} \cmidrule(lr){6-8} \cmidrule(lr){9-11}
        \textbf{Method} & \textbf{Model}
        & \textbf{FID}$\downarrow$ & \textbf{Pre.}$\uparrow$ & \textbf{Rec.}$\uparrow$
        & \textbf{FID}$\downarrow$ & \textbf{Pre.}$\uparrow$ & \textbf{Rec.}$\uparrow$
        & \textbf{FID}$\downarrow$ & \textbf{Pre.}$\uparrow$ & \textbf{Rec.}$\uparrow$ \\
        \midrule

        \multirow[c]{4}{*}{GANs}
         & StyleGAN2~\cite{karras2020analyzing}     & 179.00 & 0.411 & 0.005 & 100.16 & 0.492 & 0.156 & 54.30 & 0.483 & 0.209 \\
         & StyleGAN2+ADA~\cite{karras2020training}  & 85.80   & 0.576 & 0.007 & 21.29 & 0.689 &  0.235 &15.39  & 0.731 & 0.308 \\
         & StyleGAN2+DiffAug~\cite{zhao2020differentiable}  & 61.91 & 0.635 & 0.007 &  25.66 & 0.710 & 0.240 &  24.32 & 0.704 & 0.235 \\
         & FakeCLR~\cite{li2022fakeclr}       & 42.56 & 0.840 & 0.007 &  \textbf{15.92} & 0.727 & 0.242 &  \textbf{9.90} & \underline{0.748} & 0.300 \\
        \midrule  

        \multirow[c]{6}{*}{Diffusion}
         & EDM~\cite{karras2022elucidating}          & 79.10 & 0.512 & 0.005 & 44.19 & 0.643 & 0.194 & 31.28 & 0.683 & 0.287 \\
         & EDM + DA~\cite{karras2022elucidating}          & 50.73 & 0.724 & 0.005 &  30.75 & 0.668 & 0.222 &  27.17 & 0.709 & 0.281 \\
         & Patch Diffusion~\cite{wang2023patch}          & 44.45 & 0.746 & 0.011 &  28.03 & 0.707 & 0.224 &  25.32 & 0.718 & 0.292 \\
         & LD-Diffusion~\cite{zhang2025training}      & \textbf{28.51} & 0.791 & \underline{0.018} &  \underline{17.76} & \textbf{0.767} & \underline{0.249} &  \underline{14.36} & \textbf{0.769} & \underline{0.323} \\
         & SiT-B/2~\cite{ma2024sit}      & 43.76 & \textbf{0.854} & 0.008 &  29.89 & 0.724 & 0.189 &  26.13 & \underline{0.748} & 0.245 \\
         & SiT-B/2 (DiNO-VAE)~\cite{xu2025exploring}      & 50.12 & 0.803 & 0.006 &  33.27 & 0.710 & 0.175 &  30.00 & 0.709 & 0.237 \\
         \rowcolor{graybg}
         & \textbf{LIRF (Ours)} & \underline{34.98} & \underline{0.851} & \textbf{0.021}
                              & \textbf{20.16} & \underline{0.741} & \textbf{0.267}
                              & 15.97 & 0.734 & \textbf{0.326} \\
        \bottomrule
    \end{tabular}
    \vspace{-0.1in}
\end{table*}

\begin{definition}[\textbf{Correction operator}]\label{def:correction_operator}
Given a candidate set $\tilde{Z}^{(r)}$ at round $r$, for each generated $z_{\gen} \in \tilde{Z}^{(r)}$, we identify its top-$k$ neighbors under cosine similarity, denoted by $\mathcal{N}_k(z_{\gen})\subset Z^{(r)}$.
A local reference point $z_{\rref}$ is then constructed by aggregating this neighborhood under spherical geometry as
\begin{equation} 
z_{\rref} = \bar{r} \cdot \mathbf{u},~\text{where}~\mathbf{u} = \frac{\sum w_j z_j}{\left\| \sum w_j z_j \right\|_2},~\bar{r} = \sum w_j \|z_j\|_2,
\end{equation}
where $w_j=\mathrm{Softmax}_{z_j\in\mathcal{N}_k(z_{\gen})}(\cos(z_{\gen},z_j))$ are cosine-similarity weights that emphasize semantically closer neighbors. 
Here, $\mathbf{u}$ represents the dominant semantic direction of the local neighborhood, while $\bar{r}$ corresponds to reference radial magnitude of the local data manifold.
Then, the corrected sample $\mathcal{C}(z_{\gen})$ is obtained by contracting the candidate $z_{\gen}$ towards this local reference via spherical linear interpolation (SLERP)~\cite{white2016sampling}:
\begin{equation*}
\mathcal{C}(z_{\gen}) = \left( (1-\lambda)\|z_{\gen}\|_2 + \lambda \bar{r} \right) \cdot\text{SLERP}\left(\hat{z}_{\gen} , \mathbf{u}; \lambda\right).
\end{equation*}
where $\hat{z}_{\gen} = \frac{z_{\gen}}{\|z_{\gen}\|_2}$. 

The SLERP operator is defined, for unit vectors $a,b$ and interpolation factor $\lambda \in[0,1]$ as
\begin{equation}
\text{SLERP}(a,b;\lambda) = \frac{\sin((1-\lambda)\theta)}{\sin\theta}a + \frac{\sin(\lambda\theta)}{\sin\theta}b,
\end{equation}
where $\theta=\arccos\left(a^\top b\right)$.
\end{definition}

Intuitively, $z_{\gen}$ is an exploratory latent may lie in off-manifold regions, whereas $z_{\rref}$ is computed from nearby samples in $\mathcal{N}_k(z_{\gen})$ and thus serves as a local manifold reference.
The SLERP term interpolates between the directions of $z_{\gen}$ and $\mathbf{u}$ along the spherical geodesic, while the prefactor $(1-\lambda)\|z_{\gen}\|_2+\lambda\bar r$ simultaneously interpolates the radial scale toward the neighborhood's radius.
Overall, $\mathcal{C}(\cdot)$ pulls $z_{\gen}$ toward $z_{\rref}$ while respecting the cosine-aligned geometry of the latent space.

In the following result, we show that the correction operator $\mathcal{C}(\cdot)$ in Definition~\ref{def:correction_operator} admits a local Euclidean contraction, and consequently pushes generated samples closer to the local reference point.

\begin{proposition}[\textbf{Local Euclidean contraction of $\mathcal{C(\cdot)}$}]
Conditioning on a fixed neighborhood $\mathcal{N}_k$ and associated weights $\{w_j\}$ (and hence a fixed local reference point $p \triangleq z_{\rref}\neq 0$), the correction operator $\mathcal{C}(\cdot)$ in Definition~\ref{def:correction_operator} exhibits a local Euclidean contraction.
Specifically, there exists a tubular set $\mathcal{D}_\rho$ around $p$ and a constant $\kappa\in(0,1)$ such that for all $z\in\mathcal{D}_\rho$,
\begin{equation}
\|\mathcal{C}(z)-p\|_2 \le \kappa\,\|z-p\|_2.
\label{eq:local_euclidean_contraction}
\end{equation}
A constructive bound on $\kappa$, together with a detailed proof, is provided in \Cref{app:contraction_proof} of the appendix.
\label{prop:contraction}
\end{proposition}


\paragraph{Step 3: Augmentation.} 
Admitting all corrected samples indiscriminately may lead to error accumulation.
In particular, heavily corrected candidates often come from low-density or off-manifold regions where neighborhood estimates are less reliable, making the correction $\mathcal{C}(z)$ still noisy. 
To prevent such error amplification, we further filter corrected candidates based on the correction magnitude $\delta(z)= \| z-\mathcal{C}(z)\|_2$.
Motivated by \Cref{prop:contraction}, which characterizes $\mathcal{C}(\cdot)$ as a local contraction under stable neighborhoods, we use $\delta(z)$ as a practical indicator of whether a candidate lies within this reliable regime.

Concretely, we admit a corrected sample only if $\delta(z)\le\tau$, forming the accepted set
\begin{equation}\label{eq:def_tau}
\mathcal{A}^{(r)} \triangleq \{\, \mathcal{C}(z)\;:\; z\in \tilde{Z}^{(r)},\ \delta(z)\le \tau \,\}
\end{equation}
and update the training set as $Z^{(r+1)}=Z^{(r)}\cup \mathcal{A}^{(r)}$.
This selective augmentation retains locally consistent samples while discarding heavily corrected outliers, thereby stabilizing the manifold densification process.

In the following, we establish convergence properties of the resulting three-step iterative manifold densification procedure; the proof is given in \Cref{app:densification} of the appendix. 

\begin{theorem}[\textbf{Convergence of manifold densification}]\label{thm:densification}
Let $\mathcal{M}_{Z}\subset\mathbb{R}^d$ be a compact $C^2$ submanifold of intrinsic dimension $d_m$. 
Let $\{Z^{(r)}\}_{r\ge 0}$ denote the training sets produced by LIRF, and define
$N_r \triangleq |Z^{(0)}| + m_{\mathrm{eff}}\,r$.
Under the regularity conditions stated in \Cref{app:densification}, there exist constants $C_1,C_2,C_3>0$ such that for all $r\ge 0$,
{
\small
\begin{equation}
d_H\!\big(Z^{(r)},\mathcal{M}_{Z}\big)
\;\le\;
C_1\,\kappa^{r}\, d_H\!\big(Z^{(0)},\mathcal{M}_{Z}\big)
\;+\;
C_2N_r^{-1/d_m}+C_3\,\tau,
\end{equation}
}

where $\kappa\in(0,1)$ is the local contraction factor in \Cref{prop:contraction}, $\tau$ is the admission threshold in Eq.~\eqref{eq:def_tau}, and $m_{\mathrm{eff}}=\min_r|\mathcal{A}^{(r)}|$ denotes the minimum number of admitted samples per round.
\end{theorem}

\begin{table*}[t]
    \centering
    \small
    \caption{Comparison of generation quality and diversity on Low-Shot datasets using $50$K generated samples. We report FID ($\downarrow$), Precision ($\uparrow$), and Recall ($\uparrow$). The best results are shown in \textbf{bold}, and the second-best results are \underline{underlined}. }
    \label{tab:low_shot_results}
    \setlength{\tabcolsep}{3.5pt}
    \renewcommand{\arraystretch}{1.1}
    \begin{tabular}{@{}l p{5.6cm} ccc ccc ccc@{}}
        \toprule
        & & \multicolumn{3}{c}{\textbf{Obama}} & \multicolumn{3}{c}{\textbf{Grumpy Cat}} & \multicolumn{3}{c}{\textbf{Panda}} \\
        \cmidrule(lr){3-5} \cmidrule(lr){6-8} \cmidrule(lr){9-11}
        \textbf{Method} & \textbf{Model}
        & \textbf{FID}$\downarrow$ & \textbf{Pre.}$\uparrow$ & \textbf{Rec.}$\uparrow$
        & \textbf{FID}$\downarrow$ & \textbf{Pre.}$\uparrow$ & \textbf{Rec.}$\uparrow$
        & \textbf{FID}$\downarrow$ & \textbf{Pre.}$\uparrow$ & \textbf{Rec.}$\uparrow$ \\
        \midrule
        \multirow{4}{*}{GANs}
        & StyleGAN2~\cite{karras2020analyzing}
            & 80.20 & 0.413 & 0.159  & 48.90 & 0.492 & 0.187  & 34.27 & 0.501 & 0.174 \\
        & StyleGAN2+ADA~\cite{karras2020training}
            & 45.69 & 0.577 & 0.200  & 26.62 & 0.680 & 0.237  & 12.90 & 0.731 & 0.303 \\
        & StyleGAN2+DiffAug~\cite{zhao2020differentiable}
            & 46.87 & 0.632 & 0.175  & 27.08 & 0.711 & 0.243  & 12.06 & 0.765 & 0.238 \\
        & FakeCLR~\cite{li2022fakeclr}
            & 26.95 & 0.757 & 0.305  & \underline{19.56} & 0.876 & \underline{0.417}  & \underline{8.42}  & 0.843 & 0.375 \\
        \midrule
        \multirow{6}{*}{Diffusion}
        & EDM~\cite{karras2022elucidating}
            & 51.30 & 0.947 & 0.211  & 36.90 & 0.924 & 0.298  & 23.70 & 0.905 & 0.208 \\
        & EDM + DA~\cite{karras2022elucidating}
            & 37.10 & 0.965 & 0.380  & 29.94 & 0.870 & 0.330  & 10.81 & 0.874 & 0.360 \\
        & Patch Diffusion~\cite{wang2023patch}
            & 41.47 & 0.952 & \underline{0.410}  & 30.89 & 0.861 & 0.370  & 13.25 & 0.866 & \underline{0.392} \\
        & LD-Diffusion~\cite{zhang2025training}
            & \textbf{13.00} & \textbf{0.996} & 0.100
            & \textbf{13.31} & \textbf{0.994} & 0.080
            & \textbf{4.70}  & \textbf{1.000} & 0.012 \\
        & SiT-B/2~\cite{ma2024sit}
            & 38.24 & \underline{0.977} & 0.157
            & 34.19 & \underline{0.974} & 0.111
            & 32.96 & \underline{0.967} & 0.251 \\
        & SiT-B/2 (DiNO-VAE)~\cite{xu2025exploring}
            & 40.98 & 0.833 & 0.160
            & 41.36 & 0.894 & 0.193
            & 36.31 & 0.867 & 0.204\\
        \rowcolor{graybg}
        & \textbf{LIRF (Ours)}
            & \underline{24.11} & 0.871 & \textbf{0.671}
            & 22.03 & 0.867 & \textbf{0.531}
            & 11.95 & 0.822 & \textbf{0.692} \\
        \bottomrule
    \end{tabular}
    \vspace{-0.1in}
\end{table*}

\section{Experiments}

In this section, we evaluate LIRF on FFHQ \cite{karras2020analyzing} and the Low-Shot datasets \cite{zhao2020differentiable}, following prior work~\cite{karras2020training,zhao2020differentiable,li2022fakeclr,zhang2025training}. 
For fair comparison, all images are resized to $256\times256$ following \citet{zhao2020differentiable}. 
The Low-Shot benchmark consists three datasets (Obama, Grumpy Cat, Panda), each containing $100$ training images. 
The FFHQ dataset contains $\sim70$K high-resolution face images. 
Following common experimental settings~\cite{li2022fakeclr,zhang2025training}, we construct training subsets of size $N \in \{100, 1\mathrm{k}, 2\mathrm{k}\}$ from FFHQ.
As the generative backbone, we adopt the Scalable Interpolant Transformer (SiT)~\cite{ma2024sit} with the SiT-B/2 configuration, which provides a favorable balance between model capacity and data availability. 
To enable the proposed geometric-aware correction, we encode images into a semantically aligned latent space using a pretrained DiNO-VAE~\cite{xu2025exploring} (with downsampling factor $8$) as discussed in \Cref{sub:semantically_aligned_latent_space}. 
This replaces the SD-VAE~\cite{Rombach_2022_CVPR} commonly used in standard flow-matching pipelines. For the baselines, we strictly follow the reported numbers from existing studies using identical datasets and evaluation metrics ~\cite{karras2020training,zhao2020differentiable,zhang2025training}.

\textbf{Training Settings.} 
For LIRF, we linearly anneal the interpolation factor $\lambda$ from $0.8$ to $0.2$ over the course of training and fix the admission threshold at $\tau=0.1$. 
Sampling is performed using the \texttt{dopri5} ODE solver with $250$ function evaluations (NFE), consistent with the default SiT setup. 
All remaining hyperparameters and architectural choices follow the default SiT settings~\cite{ma2024sit}. 
Implementation details are provided in \Cref{app:hyperparameters} of the appendix.

\subsection{Results on FFHQ}

We evaluate LIRF on FFHQ subsets at $256\times256$ resolution with $N \in \{100, 1\mathrm{k}, 2\mathrm{k}\}$. 
\Cref{tab:ffhq_results} reports FID computed from $50$K generated samples, together with Precision and Recall~\cite{kynkaanniemi2019improved}, evaluated against the full FFHQ dataset containing $70$K images.
Compared with the vanilla SiT backbone, LIRF consistently improves both FID and coverage metrics across all data regimes. 
On FFHQ-100 subset, LIRF reduces FID from $43.76$ to $34.98$ and increases the Recall from $0.008$ to $0.021$, indicating substantially improved coverage beyond the sparse training samples. 
While LD-Diffusion achieves the lowest FID, LIRF attains the \emph{highest Recall} across all subset sizes.

\subsection{Results on Low-Shot Datasets}
We further evaluate LIRF on the Low-Shot benchmark~\cite{zhao2020differentiable}, including the Obama, Grumpy Cat, and Panda subsets, each containing $100$ training images.
\Cref{tab:low_shot_results} reports FID computed from $50$K generated samples, together with Precision and Recall evaluated against the $100$ training images, following~\cite{zhao2020differentiable}.
Compared with the vanilla SiT-B/2 backbone, LIRF consistently reduces FID across all three datasets and substantially improves Recall. 

Relative to diffusion-based baselines, LIRF achieves \emph{notably higher coverage} as reflected by Recall. 
While LD-Diffusion attains the lowest FID and the highest Precision all datasets, its substantially lower Recall suggests a tendency towards overfitting.
In contrast, LIRF achieves the highest Recall on all three datasets, outperforming the second-best method by a large margin, with \emph{competitive} FID and Precision.
These results are consistent with the goal of our geometry-aware refinement: improving global coverage under data scarcity while maintaining competitive sample quality.

\subsection{Ablation Study}

In this section, we conduct ablation studies on the FFHQ-100 subset to examine the contribution of individual components in LIRF. 
We first compare DiNO-VAE with a standard SD-VAE to evaluate the importance of a semantically aligned latent space. 
And then analyze the sensitivity of LIRF to the correction factor $\lambda$ and the admission threshold $\tau$. 
Next, we investigate the trade-off between generation quality and computational cost controlled by the refinement interval $\Delta$. 
Finally, we study the effect of the candidate pool size on the effectiveness of latent manifold densification.

\begin{table}[t]
    \centering
    \footnotesize
    \caption{\textbf{Effectiveness of semantically aligned Latent Space}. Comparison of LIRF performance when using a standard SD-VAE versus DiNO-VAE, with all other hyperparameters held constant.}
    \label{tab:ablation_vae}
    \begin{tabular}{lccc}
        \toprule
        \small
        \textbf{Latent Configuration} & \textbf{FID} $\downarrow$ & \textbf{Pre.} $\uparrow$ & \textbf{Rec.} $\uparrow$ \\
        \midrule
        LIRF SD-VAE & 42.15 & \textbf{0.883} & 0.009 \\
        LIRF DiNO-VAE (\textbf{Ours}) & \textbf{34.98} & 0.851 & \textbf{0.021} \\
        \bottomrule
    \end{tabular}
    \vspace{-0.1in}
\end{table}
\paragraph{Effectiveness of semantically aligned latent space.}
\Cref{tab:ablation_vae} compares LIRF built on SD-VAE \cite{Rombach_2022_CVPR} and with DiNO-VAE~\cite{xu2025exploring}. 
When combined with SD-VAE, LIRF yields performance close to the vanilla SiT baseline, with only marginal improvements. 
This is likely due to the lack of local semantic structure in SD-VAE, where the correction operator $\mathcal{C}(\cdot)$ frequently traverses low-density regions, producing artifacts that fail the admission criterion.

\begin{figure}[ht]
    \centering
    \includegraphics[width=0.9\columnwidth]{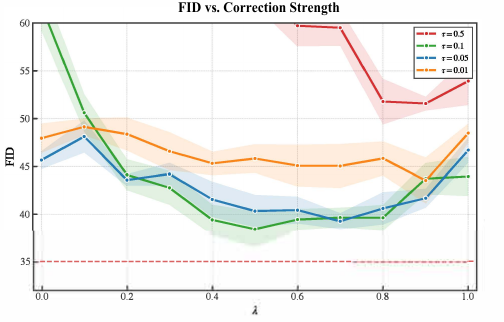}
    \caption{\textbf{Sensitivity analysis of correction strength $\lambda$ and admission threshold $\tau$ on FFHQ-100}. The curves show FID for different fixed values of $\lambda$ under varying $\tau$. 
    Shaded regions indicate the standard deviation over 5 runs. 
    The dashed line corresponds to the adaptive schedule, where $\lambda$ is linearly decayed from $0.8$ to $0.2$.}
    \label{fig:ablation_sensitivity}
    \vspace{-0.1in}
\end{figure}

\paragraph{Sensitivity to correction strength $(\lambda, \tau)$.}
We analyze the interaction between the admission threshold $\tau$ and the correction strength $\lambda$. 
As shown in \Cref{fig:ablation_sensitivity}, $\tau$ primarily acts as a quality filter: a loose threshold admits distorted candidates and may introduce noise, whereas an overly strict threshold limits manifold expansion. 
In our experiments, $\tau = 0.1$ provides the best trade-off.
For the correction strength, fixed $\lambda$ induce a trade-off between geometric fidelity and generative diversity. 
We therefore adopt a linear decay schedule for $\lambda$ from $0.8$ to $0.2$ over training. 
This schedule enforces stronger correction in early stages to suppress off-manifold artifacts, and gradually relaxes the constraint to preserve diversity as the model improves.

\begin{table}[t]
    \centering
    \scriptsize
    \caption{\textbf{Effect of refinement frequency $\Delta$}. Comparison of generation quality (FID) and training cost (minutes per $10$k steps) across different refinement intervals.}
    \label{tab:ablation_frequency}
    \begin{tabular}{l cccc}
        \toprule
        \textbf{Metric} & \textbf{$\Delta=10\text{k}$} &  \textbf{$\Delta=50\text{k}$} & \textbf{$\Delta=100\text{k}$} & \textbf{Baseline} \\
        \midrule
        \textbf{FID} $\downarrow$ & 35.42 & 34.98 & 44.71 & 43.76 \\
        \textbf{Time (mins)} &  99& 91 & 90 & 89 \\
        \bottomrule
    \end{tabular}
    \vspace{-0.1in}
\end{table}
\paragraph{Effect of refinement frequency $\Delta$.}

We next evaluate the impact of the refinement interval $\Delta$.  
As shown in \Cref{tab:ablation_frequency}, increasing the interval from $\Delta=10\mathrm{k}$ to $\Delta=50\mathrm{k}$ slightly improves FID, while reducing the training time per $10$K steps from $99$ to $91$ minutes. 
However, further increasing to $\Delta=100\mathrm{k}$ leads to a noticeable performance degradation, even below the baseline.
These results suggest that refinement must be performed sufficiently frequently. 
When $\Delta$ is too large, the model may overfit sparse training samples before being properly augmented. 
Based on this trade-off, we use $\Delta=50\mathrm{k}$ as the default setting, balancing generation quality and computational efficiency.

\paragraph{{Effect of candidate set size $M$}.}

We further study the effect of the candidate set size by varying the ratio $M/N\in\{30\%,50\%,100\%,150\%\}$. 
As shown in \Cref{tab:ablation_size}, smaller candidate pools ($30\%$ or $50\%$) yield too few admitted samples to effectively densify the latent manifold, leaving the model prone to velocity field collapse. 
Setting $M=100\%\,N$ provides the best trade-off (FID $=34.98$ with improved Recall), while $150\%N$ leads to only marginal changes, indicating saturation. 
We therefore adopt $M = 100\%\,N$ as the default setting.

\begin{table}[t]
    \centering
    \footnotesize
    \caption{\textbf{Effect of candidate set size $M$}. FID, Precision, and Recall on FFHQ-100 with varying candidate set sizes $M$ relative to the training data size ($N$).}
    \label{tab:ablation_size}
    \begin{tabular}{l c c c c}
        \toprule
        \textbf{Ratio \tiny{($M/N$)}} & \textbf{30\%} & \textbf{50\%} & \textbf{100\%} & \textbf{150\%} \\
        \midrule
        \textbf{FID} $\downarrow$ & 41.52 & 38.15 & 34.98 & 35.13 \\
        \textbf{Pre.} $\uparrow$ & 0.861 & 0.858 & 0.851 & 0.849 \\
        \textbf{Rec.} $\uparrow$ & 0.010 & 0.013 & 0.021 & 0.021 \\
        \bottomrule
    \end{tabular}
    \vskip -0.1in
\end{table}

\section{Conclusion}

We identified \emph{velocity field collapse} as an important cause of memorization in generative models under data scarcity. 
To address this, we propose Latent Iterative Refinement Flow (LIRF), which progressively densifies the data manifold via a \emph{generation--correction--augmentation} loop. 
LIRF restores manifold continuity and significantly outperforms diffusion baselines on FFHQ and Low-Shot benchmarks.

Furthermore, we aim to extend LIRF to high-resolution synthesis and diverse domains. Future research will focus on developing adaptive refinement schedules based on training statistics and conducting finer-grained cost analyses to establish scalable defaults.







\section*{Impact Statement}

This paper improves the reliability of diffusion/flow-matching models in the limited-data regime by mitigating collapse-to-memorization through geometry-aware latent manifold densification. A positive impact is enabling higher-diversity generation when large datasets are unavailable, which can benefit data-scarce scientific and domain-specific applications and may reduce training-sample reproduction. However, making generative modeling more data-efficient can also lower barriers for misuse (e.g., generating deceptive imagery), and limited datasets may encode biases that could be amplified. We therefore recommend auditing for memorization/privacy leakage and bias on the target data, and adopting appropriate downstream safeguards when releasing or deploying models.


\bibliography{LIRF_short}
\bibliographystyle{icml2026}


\clearpage
\appendix
\section{Theoretical Analysis}
\subsection{Local Euclidean Contraction of the Correction Operator}
\label{app:contraction_proof}
\paragraph{Setup.} 
Fix a reference neighborhood $\mathcal N_k$ and weights $\{w_j\}$, which induces a fixed local reference point
\begin{equation}
\label{eq:app:def_p_ref}
\small
p \triangleq z_{\mathrm{ref}}=\bar r\,\mathbf u \neq 0,
\quad
\mathbf u=\frac{\sum_j w_j z_j}{\big\|\sum_j w_j z_j\big\|_2},
\quad
\bar r=\sum_j w_j\|z_j\|_2 .
\end{equation}
Write polar decompositions
$
\small
z = r s, r=\|z\|_2,\ s=\frac{z}{\|z\|_2}\in\mathbb S^{d-1}$,
and the reference point 
$p = R u, R=\|p\|_2,\ u=\frac{p}{\|p\|_2}\in\mathbb S^{d-1}.$
Conditioned on $p$, define the correction operator $C(z)=r'(z)\,s'(z),$
\begin{equation}
\label{eq:app:def_Cp}
\small
r'(z)=(1-\lambda)r+\lambda R,
\quad
s'(z)=\mathrm{SLERP}(s,u;\lambda).
\end{equation}
where $\lambda\in(0,1)$. Note that $\mathcal C(p)=p$.
When $s$ and $u$ are antipodal ($\angle(s,u)=\pi$), the shortest geodesic is not unique.
In practice, $\mathrm{SLERP}$ is implemented with a consistent convention\cite{white2016sampling}. Throughout, statements involving $\mathrm{SLERP}$ should be understood under such a consistent choice.
\begin{lemma}[Angular shrinkage of SLERP]
\label{lem:app:slerp_angle}
Let $\theta=\angle(s,u)\in[0,\pi]$. Then
\begin{equation}
\label{eq:app:slerp_angle}
\angle\!\big(\mathrm{SLERP}(s,u;\lambda),u\big)=(1-\lambda)\theta.
\end{equation}
\end{lemma}
\begin{proof}
SLERP parameterizes the shortest geodesic on $\mathbb S^{d-1}$ from $s$ to $u$ at constant speed.
Hence the remaining geodesic distance to $u$ is the remaining fraction $1-\lambda$ of the original distance.
\end{proof}
\begin{lemma}[Cosine-gap contraction on a bounded angle range]
\label{lem:app:cos_gap}
Fix $\theta_{\max}\in(0,\pi]$, define $a\triangleq 1-\lambda\in(0,1)$, and
\begin{equation}
\label{eq:app:def_cang}
c_{\mathrm{ang}}(a;\theta_{\max})
\triangleq
\sup_{\theta\in(0,\theta_{\max}]}\frac{1-\cos(a\theta)}{1-\cos\theta}.
\end{equation}
Then $c_{\mathrm{ang}}(a;\theta_{\max})\in(0,1)$ and for all $\theta\in[0,\theta_{\max}]$,
\begin{equation}
\label{eq:app:cos_gap_contract}
1-\cos(a\theta)\ \le\ c_{\mathrm{ang}}(a;\theta_{\max})\,(1-\cos\theta).
\end{equation}
\end{lemma}
\begin{proof}
For any $\theta\in(0,\theta_{\max}]$, since $a\theta<\theta$ and $\cos$ is strictly decreasing on $[0,\pi]$,
we have $\cos(a\theta)>\cos\theta$, hence the ratio in \eqref{eq:app:def_cang} is strictly $<1$.

To justify that the supremum is still $<1$, note the ratio admits a continuous extension at $\theta=0$:
by Taylor expansion,
\[
\frac{1-\cos(a\theta)}{1-\cos\theta}
\;\xrightarrow[\theta\to 0]{}\;
a^2 < 1.
\]
Therefore the ratio is bounded away from $1$ on the compact interval $[0,\theta_{\max}]$,
implying $c_{\mathrm{ang}}(a;\theta_{\max})\in(0,1)$ and \eqref{eq:app:cos_gap_contract}.
\end{proof}

\paragraph{Local Euclidean contraction}
\begin{proposition}[Local Euclidean contraction of $\mathcal C(\cdot)$]
\label{prop:app:local_contraction}
Define the local domain
\begin{equation}
\label{eq:app:def_D_rho}
\mathcal D_{\rho,\theta_{\max}}(p)
\triangleq
\left\{z\neq 0:\ \frac{R}{\rho}\le \|z\|_2 \le \rho R,\ \angle\!\left(\frac{z}{\|z\|_2},u\right)\le \theta_{\max}\right\},
\end{equation}
where $\rho\ge 1,\ \theta_{\max}\in(0,\pi].$

Let $a=1-\lambda$ and $c_{\mathrm{ang}}(a;\theta_{\max})$ be defined in \eqref{eq:app:def_cang}.
Then for all $z\in\mathcal D_{\rho,\theta_{\max}}(p)$,
\begin{equation}
\label{eq:app:local_euclidean_contraction}
\|\mathcal C(z)-p\|_2 \le \kappa\,\|z-p\|_2,
\end{equation}
where a valid contraction factor is
\begin{equation}
\label{eq:app:def_kappa}
\kappa
\triangleq
\max\left\{
a,\ \sqrt{\big(a+(1-a)\rho\big)\,c_{\mathrm{ang}}(a;\theta_{\max})}
\right\}.
\end{equation}
In particular, if $\big(a+(1-a)\rho\big)c_{\mathrm{ang}}(a;\theta_{\max})<1$ then $\kappa\in(0,1)$ and $\mathcal C$
is a strict Euclidean contraction towards $p$ on $\mathcal D_{\rho,\theta_{\max}}(p)$.
\end{proposition}

\begin{proof}
Let $\theta=\angle(s,u)$. A standard identity gives
\begin{equation}
\label{eq:app:zp_decomp}
\|z-p\|_2^2
=
\|rs-Ru\|_2^2
=
(r-R)^2 + 2rR(1-\cos\theta).
\end{equation}
Let $z'=\mathcal C(z)=r's'$ with $r'=ar+(1-a)R$ and $s'=\mathrm{SLERP}(s,u;\lambda)$.
By Lemma~\ref{lem:app:slerp_angle}, $\theta'=\angle(s',u)=a\theta$, hence
\begin{equation}
\label{eq:app:zprimep_decomp}
\|z'-p\|_2^2
=
(r'-R)^2 + 2r'R(1-\cos\theta').
\end{equation}
For the radial term,
\[
r'-R = a(r-R)\quad\Rightarrow\quad (r'-R)^2=a^2(r-R)^2.
\]
For the angular term, Lemma~\ref{lem:app:cos_gap} yields
$1-\cos\theta' \le c_{\mathrm{ang}}(a;\theta_{\max})(1-\cos\theta)$ for $\theta\le \theta_{\max}$.
Moreover, for $z\in\mathcal D_{\rho,\theta_{\max}}(p)$ we have $r\ge R/\rho$, hence $R/r\le \rho$ and
\[
\frac{r'}{r} = a + (1-a)\frac{R}{r}\le a+(1-a)\rho.
\]
Therefore,
\[
2r'R(1-\cos\theta')
\le
\big(a+(1-a)\rho\big)c_{\mathrm{ang}}(a;\theta_{\max})\cdot 2rR(1-\cos\theta).
\]
Combining \eqref{eq:app:zprimep_decomp} and \eqref{eq:app:zp_decomp}, letting
$b\triangleq \big(a+(1-a)\rho\big)c_{\mathrm{ang}}(a;\theta_{\max})$, we obtain
\[
\|z'-p\|_2^2
\le
a^2(r-R)^2 + b\cdot 2rR(1-\cos\theta)
\le
\max\{a^2,b\}\,\|z-p\|_2^2.
\]
Taking square roots yields \eqref{eq:app:local_euclidean_contraction} with $\kappa=\max\{a,\sqrt b\}$.
\end{proof}

\subsection{Convergence Analysis for Manifold Densification}
\label{app:densification}

Let $\mathcal M_{\mathcal Z}\subset\mathbb R^d$ be a compact $C^2$ submanifold of intrinsic dimension $d_m$.
Define $\mathrm{dist}(z,\mathcal M)\triangleq \inf_{x\in\mathcal M}\|z-x\|_2$.
Let $\mathcal U_{\rho_{\mathcal M}}\triangleq\{z:\mathrm{dist}(z,\mathcal M_{\mathcal Z})<\rho_{\mathcal M}\}$ be a tubular neighborhood
on which the nearest-point projection $\Pi:\mathcal U_{\rho_{\mathcal M}}\to\mathcal M_{\mathcal Z}$ is well-defined.
For the corrected training sets $\{\mathcal Z^{(r)}\}_{r\ge 0}$, define
\begin{equation}
\label{eq:app:def_e_h}
e_r \triangleq \sup_{z\in\mathcal Z^{(r)}} \mathrm{dist}(z,\mathcal M_{\mathcal Z}),
\quad
h_r \triangleq \sup_{x\in\mathcal M_{\mathcal Z}}\mathrm{dist}\big(x,\Pi(\mathcal Z^{(r)})\big).
\end{equation}

\subsubsection{Assumptions}
\paragraph{Assumption 1 (Tubular validity).}
There exists $\rho_{\mathcal M}>0$ such that $\mathcal Z^{(r)}\subset\mathcal U_{\rho_{\mathcal M}}$ for all $r$.

\paragraph{Assumption 2 (Local contraction; from Proposition~\ref{prop:app:local_contraction}).}
Conditioning on a fixed retrieved neighborhood and weights (hence a fixed $p\neq 0$), the correction operator $\mathcal C^{(r)}$
satisfies: there exist a set $\mathcal D(p)\subset\mathcal U_{\rho_{\mathcal M}}$ and a constant $\kappa\in(0,1)$ such that for all $z\in \mathcal D(p)$,
\begin{equation}
\label{eq:app:assump_contraction}
\|\mathcal C^{(r)}(z)-p\|_2 \le \kappa\,\|z-p\|_2.
\end{equation}

\paragraph{Assumption 3 (Reference accuracy).}
There exists $\alpha>0$ such that for all $r$ and all points $z$ to which the operator is applied, the corresponding reference point
$p^{(r)}(z)$ satisfies
\begin{equation}
\label{eq:app:assump_ref_acc}
\mathrm{dist}\big(p^{(r)}(z),\mathcal M_{\mathcal Z}\big)\le \alpha\, h_r.
\end{equation}

\paragraph{Assumption 4 (Admissible regime and coverage rate).}
The admission rule uses $\delta^{(r)}(z)=\|z-\mathcal C^{(r)}(z)\|_2$ and admits only those with
$\delta^{(r)}(z)\le \tau$, where $\tau>0$ is a fixed threshold.
Moreover, the projected set satisfies the fill-distance bound
\begin{equation}
\label{eq:app:assump_fill}
h_r \le C_{\mathcal M}\,N_r^{-1/d_m},
\end{equation}
where $N_r=|\mathcal Z^{(0)}|+m_{\mathrm{eff}}r$.

\subsubsection{Main Results}

\begin{theorem}[Convergence of manifold densification]
\label{thm:app:densification}
Let $\mathcal M_{\mathcal Z}\subset\mathbb R^d$ be a compact $C^2$ submanifold of intrinsic dimension $d_m$.
Let $\{\mathcal Z^{(r)}\}_{r\ge 0}$ be the corrected training sets and define $N_r \triangleq |\mathcal Z^{(0)}| + m_{\mathrm{eff}}\,r$.
Under Assumptions 1--4, there exist constants $C_1,C_2,C_3>0$ such that for all $r\ge 0$,
\begin{equation}
\small
\label{eq:app:densification_bound}
d_H\!\big(\mathcal Z^{(r)},\mathcal M_{\mathcal Z}\big)
\le
C_1\,\kappa^{r}\, d_H\!\big(\mathcal Z^{(0)},\mathcal M_{\mathcal Z}\big)
+
C_2 N_r^{-1/d_m}+ C_3\tau,
\end{equation}
where $\kappa\in(0,1)$ is the contraction factor in Proposition~\ref{prop:app:local_contraction} and $\tau$ is the admission threshold.
\end{theorem}

\begin{proof}
\noindent \textbf{Step 1: $d_H$ is controlled by $(e_r,h_r)$.}
By Assumption 1, $\Pi$ is well-defined on $\mathcal Z^{(r)}$.
For any $x\in\mathcal M_{\mathcal Z}$ and any $z\in\mathcal Z^{(r)}$, the triangle inequality gives
\[
\|x-z\|_2 \le \|x-\Pi(z)\|_2 + \|z-\Pi(z)\|_2.
\]
Taking $\inf_{z\in\mathcal Z^{(r)}}$ and then $\sup_{x\in\mathcal M_{\mathcal Z}}$ yields
\[
\sup_{x\in\mathcal M_{\mathcal Z}}\mathrm{dist}\big(x,\mathcal Z^{(r)}\big)
\le
h_r + e_r.
\]
Since $\sup_{z\in\mathcal Z^{(r)}}\mathrm{dist}(z,\mathcal M_{\mathcal Z})=e_r$, we obtain
\begin{equation}
\label{eq:app:hausdorff_by_eh}
d_H\big(\mathcal Z^{(r)},\mathcal M_{\mathcal Z}\big)\le h_r + e_r.
\end{equation}

\noindent \textbf{Step 2: One-step recursion for $e_{r+1}$.}
Fix any point $y$ to which $\mathcal C^{(r)}$ is applied at round $r$, and let $p=p^{(r)}(y)$ be its reference point.
Then
\begin{equation}
\label{eq:app:dist_to_M}
\mathrm{dist}\big(\mathcal C^{(r)}(y),\mathcal M_{\mathcal Z}\big)
\le
\|\mathcal C^{(r)}(y)-p\|_2 + \mathrm{dist}\big(p,\mathcal M_{\mathcal Z}\big).
\end{equation}
By Assumption 2 and the triangle inequality,
\[
\|\mathcal C^{(r)}(y)-p\|_2
\le
\kappa\|y-p\|_2
\le
\kappa\big(\mathrm{dist}(y,\mathcal M_{\mathcal Z})+\mathrm{dist}(p,\mathcal M_{\mathcal Z})\big).
\]
Combining with Assumption 3 in \eqref{eq:app:dist_to_M} gives
\begin{equation}
\label{eq:app:basic_bound}
\mathrm{dist}\big(\mathcal C^{(r)}(y),\mathcal M_{\mathcal Z}\big)
\le
\kappa\,\mathrm{dist}(y,\mathcal M_{\mathcal Z}) + (1+\kappa)\alpha\,h_r.
\end{equation}

Now consider two cases.

\emph{(i) $u\in\mathcal Z^{(r)}$.}
Then $\mathrm{dist}(u,\mathcal M_{\mathcal Z})\le e_r$, hence
\begin{equation}
\label{eq:app:old_case}
\mathrm{dist}\big(\mathcal C^{(r)}(u),\mathcal M_{\mathcal Z}\big)
\le
\kappa e_r + (1+\kappa)\alpha h_r.
\end{equation}

\emph{(ii) $y$ is admitted at round $r$.}
By the admission rule, $y=\mathcal C^{(r)}(z)$ for some candidate $z$ with $\|z-y\|_2\le \gamma_r$.
Then $\mathrm{dist}(z,\mathcal M_{\mathcal Z})\le \mathrm{dist}(y,\mathcal M_{\mathcal Z})+\gamma_r$.
Applying \eqref{eq:app:basic_bound} to $z$ and using $u=\mathcal C^{(r)}(z)$ yields
\[
\mathrm{dist}(y,\mathcal M_{\mathcal Z})
\le
\kappa\big(\mathrm{dist}(y,\mathcal M_{\mathcal Z})+\gamma_r\big) + (1+\kappa)\alpha h_r,
\]
hence
\begin{equation}
\label{eq:app:new_case}
\mathrm{dist}(y,\mathcal M_{\mathcal Z})
\le
\frac{\kappa}{1-\kappa}\gamma_r + \frac{1+\kappa}{1-\kappa}\alpha h_r.
\end{equation}

Combining \eqref{eq:app:old_case}--\eqref{eq:app:new_case} and taking supremum over $y\in\mathcal Z^{(r+1)}$ gives
\[
e_{r+1}\le \max\Big\{\kappa e_r + (1+\kappa)\alpha h_r,\ \frac{1+\kappa}{1-\kappa}\alpha h_r + \frac{\kappa}{1-\kappa}\gamma_r\Big\}.
\]
Using $\max(x,y)\le x+y$, we obtain the convenient recursion
\begin{equation}
\label{eq:app:e_rec}
e_{r+1} \le \kappa e_r + B_1 h_r + B_2 \gamma_r,
\quad
B_1\triangleq \frac{1+\kappa}{1-\kappa}\alpha,
\quad
B_2\triangleq \frac{\kappa}{1-\kappa}.
\end{equation}

\noindent \textbf{Step 3: Substitute rates for $h_r$ and the fixed threshold $\tau$.}
By Assumption~4, $h_r\le C_{\mathcal M}N_r^{-1/d_m}$ and $\gamma_r\le \tau$.
Substituting into \eqref{eq:app:e_rec} yields
\begin{equation}
\label{eq:app:e_rec2_fixedtau}
e_{r+1}
\le
\kappa e_r + B_1 C_{\mathcal M}\,N_r^{-1/d_m} + B_2 \tau.
\end{equation}

\paragraph{Step 4: Solve the recursion.}
Let $\beta \triangleq 1/d_m$ and $A \triangleq B_1 C_{\mathcal M}$.
From \eqref{eq:app:e_rec2_fixedtau} we have the recursion
\[
e_{r+1} \le \kappa e_r + A\,N_r^{-\beta} + B_2\,\tau,
\qquad \kappa\in(0,1).
\]
Unrolling yields
\begin{equation}
\label{eq:app:e_unroll_step4_fixedtau}
e_r \le \kappa^r e_0
+ A\sum_{i=0}^{r-1}\kappa^{r-1-i}N_i^{-\beta}
+ B_2\tau\sum_{i=0}^{r-1}\kappa^{r-1-i}.
\end{equation}
Re-index the first sum with $t=r-1-i$:
\[
\sum_{i=0}^{r-1}\kappa^{r-1-i}N_i^{-\beta}
=
\sum_{t=0}^{r-1}\kappa^{t}N_{r-1-t}^{-\beta}.
\]

We bound each historical term relative to $N_r^{-\beta}$:
\[
N_{r-1-t}^{-\beta}
=
N_r^{-\beta}\Big(\frac{N_r}{N_{r-1-t}}\Big)^{\beta}.
\]
Since $N_r=N_0+m_{\mathrm{eff}}r$ is nondecreasing and $N_{r-1-t}\ge N_0$ for all $t\in\{0,\ldots,r-1\}$,
\[
\frac{N_r}{N_{r-1-t}}
=
\frac{N_0+m_{\mathrm{eff}}r}{N_0+m_{\mathrm{eff}}(r-1-t)}
\le
1+\frac{m_{\mathrm{eff}}(1+t)}{N_0}.
\]
Therefore,
\[
N_{r-1-t}^{-\beta}
\le
N_r^{-\beta}\Big(1+\frac{m_{\mathrm{eff}}(1+t)}{N_0}\Big)^{\beta}.
\]
Plugging into the sum gives
\[
\sum_{t=0}^{r-1}\kappa^{t}N_{r-1-t}^{-\beta}
\le
N_r^{-\beta}\sum_{t=0}^{r-1}\kappa^t\Big(1+\frac{m_{\mathrm{eff}}(1+t)}{N_0}\Big)^{\beta}
\le
N_r^{-\beta}\,S,
\]
where
\begin{equation}
\label{eq:app:def_S}
S \triangleq \sum_{t=0}^{\infty}\kappa^t\Big(1+\frac{m_{\mathrm{eff}}(1+t)}{N_0}\Big)^{\beta} < \infty.
\end{equation}
The finiteness holds since $\kappa^t$ decays geometrically while the factor in parentheses grows only polynomially in $t$.

For the second sum in \eqref{eq:app:e_unroll_step4_fixedtau}, we use the geometric bound
\[
\sum_{i=0}^{r-1}\kappa^{r-1-i}=\sum_{t=0}^{r-1}\kappa^t \le \frac{1}{1-\kappa}.
\]
Combining the above bounds with \eqref{eq:app:e_unroll_step4_fixedtau} yields
\begin{equation}
\label{eq:app:e_final_fixedtau}
e_r \le \kappa^r e_0 + A\,S\,N_r^{-\beta} + \frac{B_2}{1-\kappa}\,\tau.
\end{equation}

\paragraph{Step 5: Conclude the Hausdorff bound.}
By \eqref{eq:app:hausdorff_by_eh}, Assumption~4 \eqref{eq:app:assump_fill}, and \eqref{eq:app:e_final_fixedtau},

\begin{equation}
\footnotesize
    d_H\big(\mathcal Z^{(r)},\mathcal M_{\mathcal Z}\big)
\le
h_r + e_r
\le
C_{\mathcal M}N_r^{-\beta}
+
\kappa^r e_0
+
A S\,N_r^{-\beta}
+
\frac{B_2}{1-\kappa}\tau.
\end{equation}

Rearranging terms and absorbing constants yields \eqref{eq:app:densification_bound} with
$C_2 \triangleq C_{\mathcal M} + AS$ and $C_3 \triangleq \frac{B_2}{1-\kappa}$.
\end{proof}

\begin{figure}[t]
    \centering
    \includegraphics[width=0.95\columnwidth]{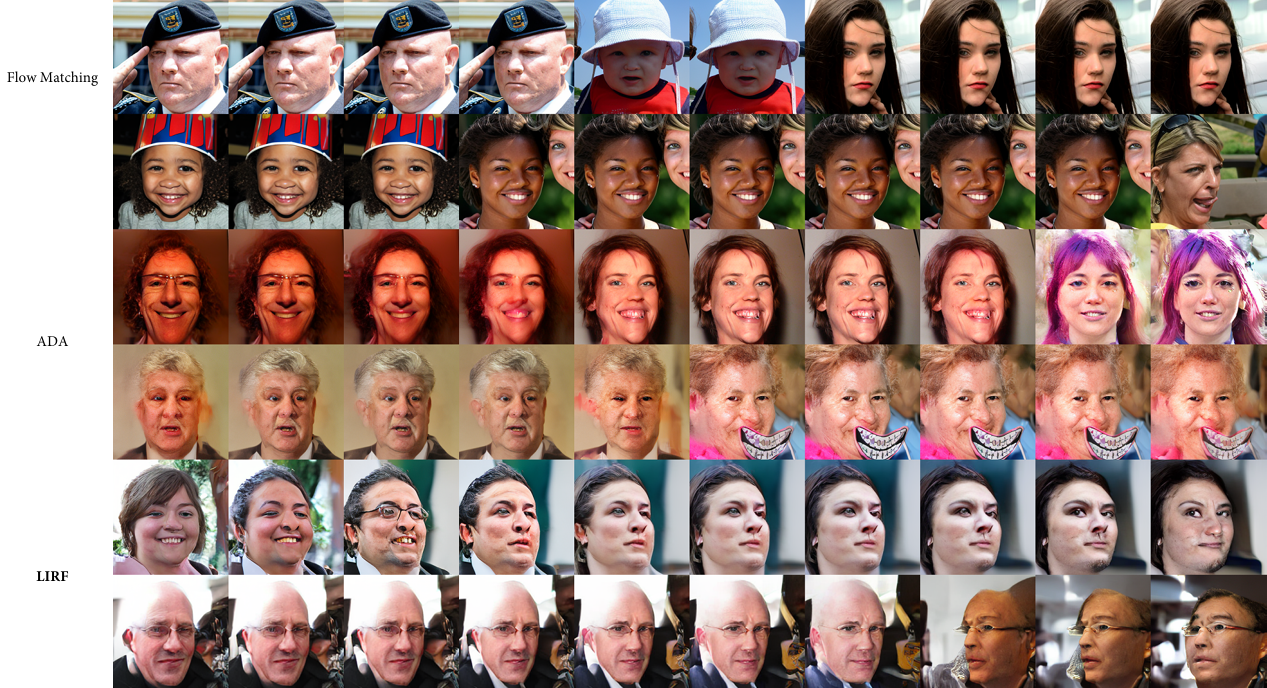}
    \caption{\textbf{Latent interpolation trajectories on FFHQ under different samplers.} For each row, we randomly sample two Gaussian noise vectors ($z^{(a)}, z^{(b)}\sim\mathcal N(0,I)$), linearly interpolate ($z(\alpha)=(1-\alpha)z^{(a)}+\alpha z^{(b)}$) with evenly spaced ($\alpha\in[0,1]$), and generate images by running the corresponding sampler from each ($z(\alpha))$.}
    \label{fig:ffhq_interp}
    \vspace{-0.1in}
\end{figure}

\begin{figure*}[t]
    \centering
    \begin{subfigure}[b]{0.23\textwidth}
        \centering
        \includegraphics[width=\linewidth, height=\linewidth]{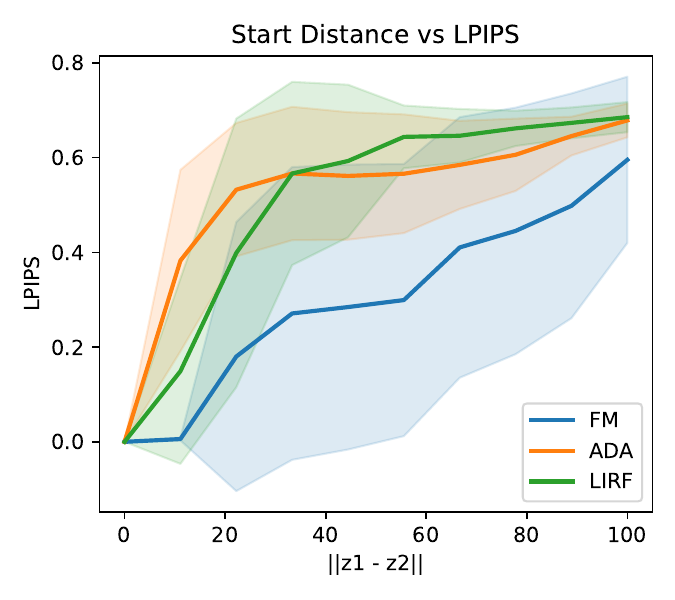} 
        \caption{Start Distance vs LPIPS}
        \label{fig:ffhq_lpips}
    \end{subfigure}
    \hfill 
    \begin{subfigure}[b]{0.74\textwidth}
        \centering
        \includegraphics[width=\linewidth, height=0.33\textwidth]{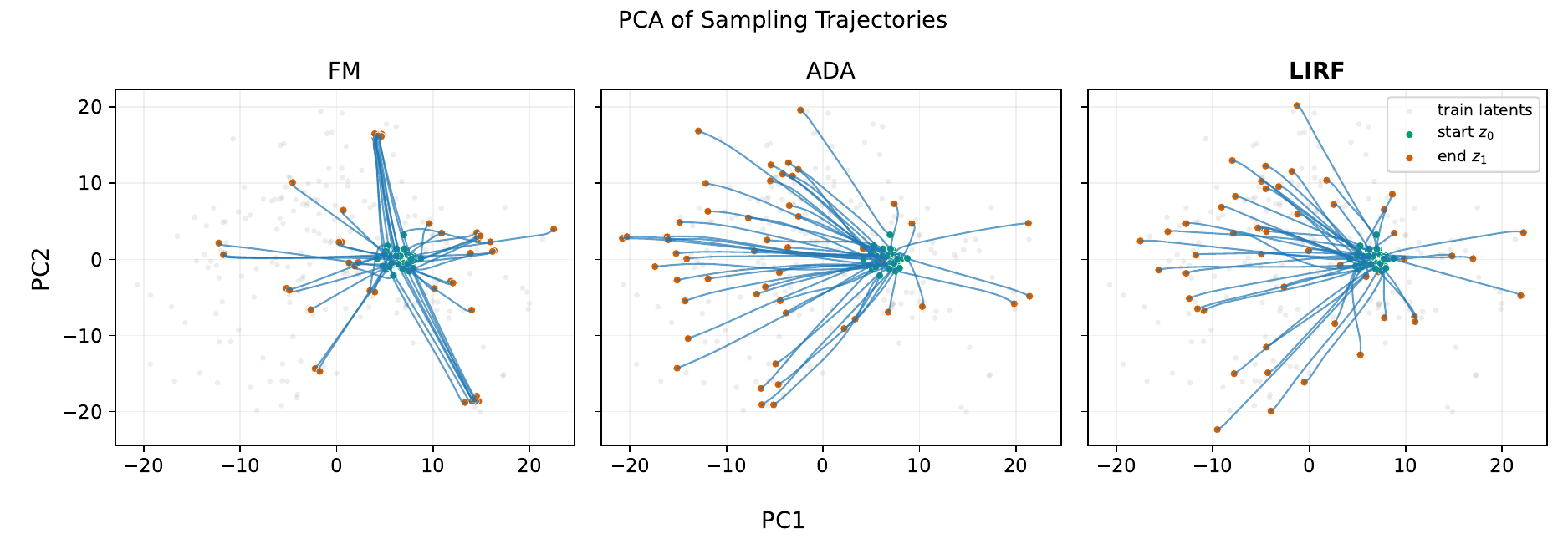} 
        \caption{Visualization of sampling trajectories}
        \label{fig:ffhq_sampling}
    \end{subfigure}
    
    \caption{\textbf{(a) Perceptual distance (LPIPS) vs. initial latent distance $\|z_1-z_2\|_2$:} results are averaged over 50 pairs of latents randomly sampled in all directions, with LPIPS computed using a VGG backbone; shaded regions represent the standard deviation (std). \textbf{(b) Visualization of sampling trajectories:} PCA projection of 50 random ODE trajectories onto the first two principal components of the latent training set.}
    \label{fig:combined_result}
\end{figure*}

\section{Experiment Details}

\subsection{Velocity field Collapse in FFHQ Datasets}
\label{app:ffhq_velocity_collapse}

We provide qualitative and quantitative evidence that, on FFHQ, the learned flow field of vanilla flow matching exhibits a pronounced \emph{velocity field collapse} towards training samples, leading to discontinuous generations under latent interpolation. We further show that LIRF substantially mitigate this behavior by yielding smoother semantic transitions and a more stable relationship between latent distance and perceptual distance.

\paragraph{Latent interpolation reveals piecewise-constant attractor basins.}
We randomly sample two Gaussian latents $z^{(a)}, z^{(b)}\sim\mathcal N(0,I)$ and generate a sequence of intermediate latents by linear interpolation
\begin{equation}
\label{eq:ffhq_interp}
z(\alpha) \triangleq (1-\alpha)z^{(a)} + \alpha z^{(b)}, \qquad \alpha\in[0,1],
\end{equation} then decode each sampler from $z(\alpha)$ to obtain images $\{x(\alpha)\}$.
Figure~\ref{fig:ffhq_interp} shows that FM produces \emph{segmented} interpolation paths: as $\alpha$ varies smoothly, the generated identity stays nearly unchanged for a range of $\alpha$ and then abruptly jumps to a different face, closely resembling a training sample. This is consistent with the latent space being partitioned into basins of attraction whose boundaries induce discontinuous changes in the terminal sample.
In contrast, ADA and LIRF exhibit more continuous, semantically meaningful transitions along $\alpha$, suggesting a less collapsed and more coherent velocity field in latent space.
 \begin{figure}[h!]
 \centering
 \includegraphics[width=0.99\columnwidth]{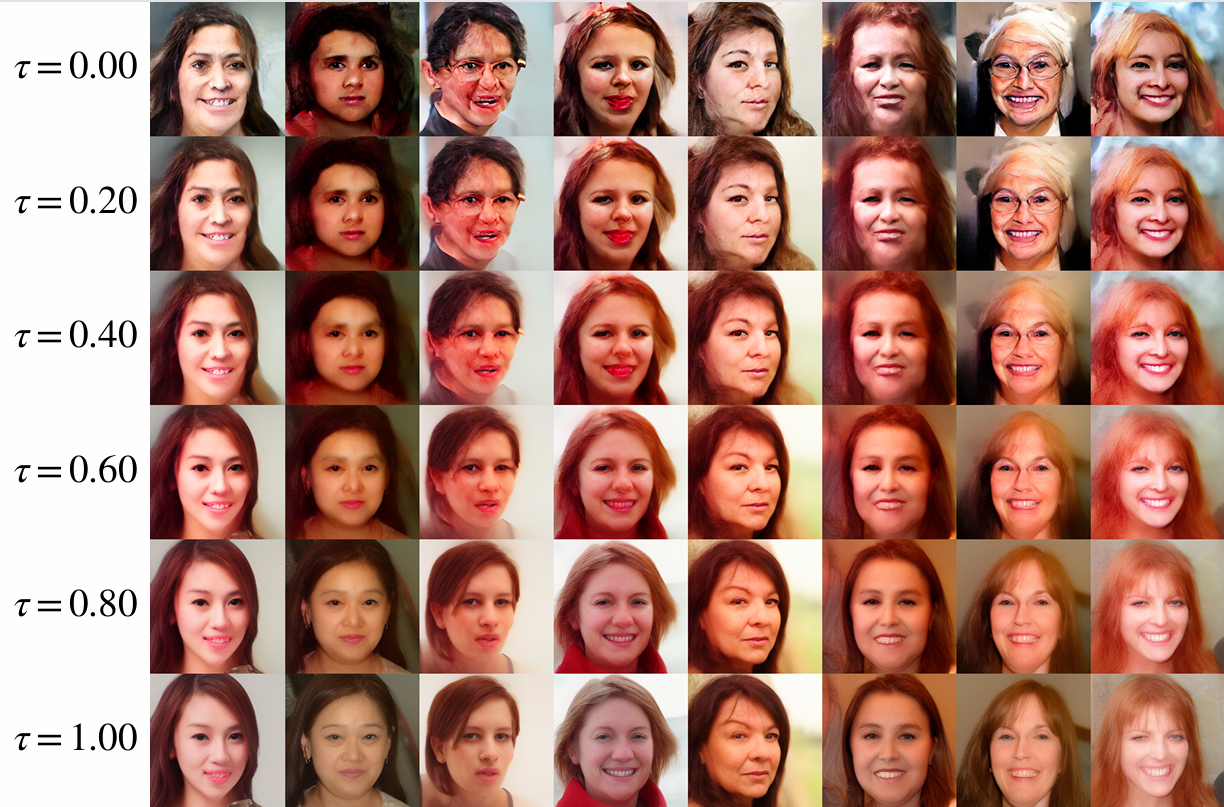}
     \caption{\textbf{Impact of hyperparameter $\lambda$ in $\mathcal{C}(\cdot)$}: At $\lambda=0$ (raw candidates), the generated samples exhibit severe structural artifacts and noise. 
     As $\lambda$ increases, the correction operator pulls these candidates towards the local manifold, progressively enhancing realism while preserving the semantic layout.}
     \label{fig:lambda_effect}
     \vspace{-0.1in}
 \end{figure}

\begin{figure*}[t]
 \centering
 \includegraphics[width=0.95\textwidth]{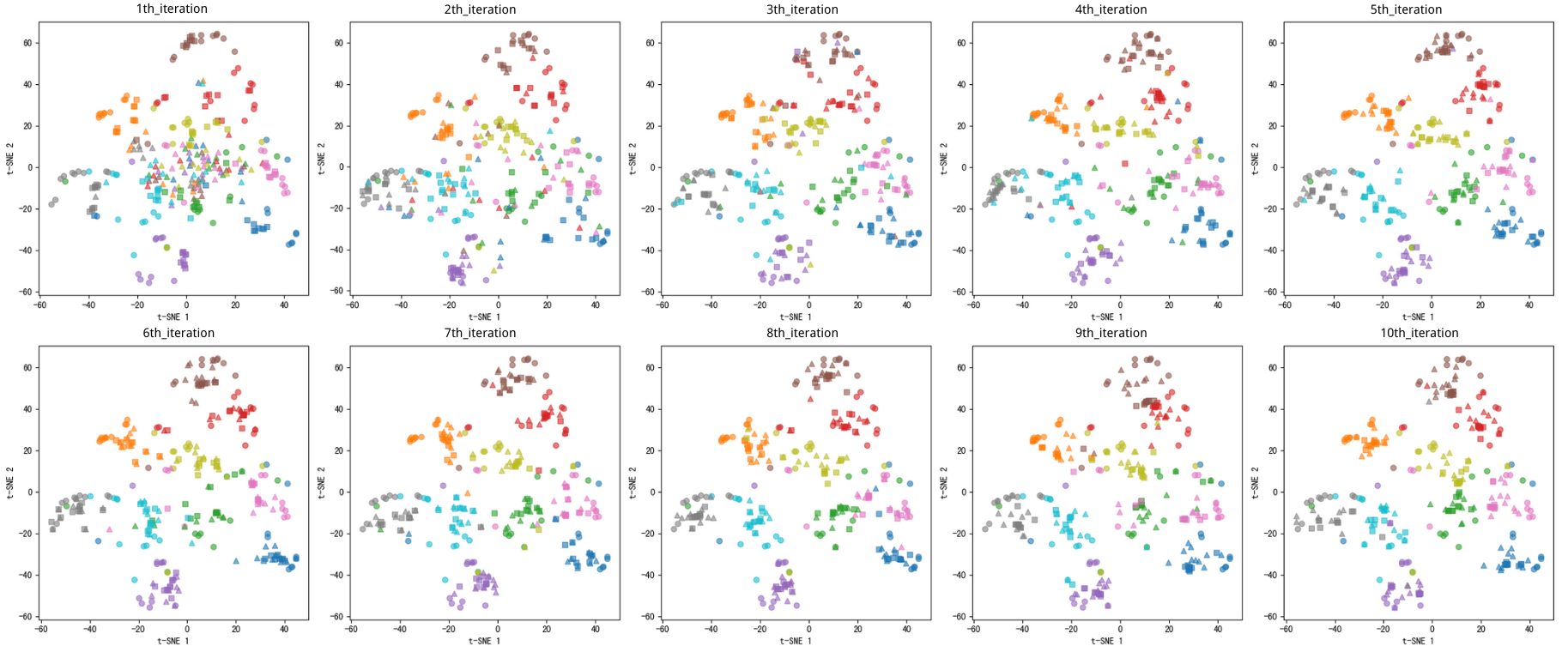}
     \caption{\textbf{Latent distribution evolution on MNIST.} Each point represents a sample: real samples are circles, generated samples are triangles, and corrected samples are squares. The results show the progression of the learned latent space across 10 iterations, highlighting the improvement in the alignment of generated and corrected samples with real data.}
     \label{fig:mnist_tsne}
     \vspace{-0.1in}
\end{figure*}

\paragraph{LPIPS vs.\ latent distance.}
To quantify the above effect, we sample pairs of initial latents $(z_1,z_2)$ and compute the perceptual distance between the resulting images using LPIPS distance,
plotting it against the latent distance $\|z_1-z_2\|_2$ (mean with variability band) in Figure~\ref{fig:ffhq_lpips}.

FM exhibits an extreme \emph{local collapse}: when $\|z_1-z_2\|_2$ is small, LPIPS is almost zero, indicating that many nearby latents are mapped to essentially the \emph{same} image (the same training sample). As $\|z_1-z_2\|_2$ increases, the variance of LPIPS under FM becomes very large, consistent with different initial latents falling into different attractor basins and being pulled towards different training samples.
By comparison, ADA and LIRF show a more monotone and stable dependence of LPIPS on $\|z_1-z_2\|_2$, and their variability decreases as $\|z_1-z_2\|_2$ grows, indicating that the mapping is less dominated by discrete attractors and more governed by gradual semantic change.

\paragraph{Visualization of sampling trajectories via PCA.}
To further expose the underlying transport dynamics, we visualize the ODE sampling trajectories by projecting the latent paths onto the first two principal components (PCs) of the training set in Figure\cref{fig:ffhq_sampling}. Under vanilla flow matching (FM), the trajectories exhibit a high degree of collapse, where multiple paths originating from distinct initial noise vectors $z_0$ (green dots) converge into nearly identical end points (orange dots). These terminal points align tightly with sparse training latents (light gray dots), providing direct visual evidence of the point attractors that define velocity field collapse.

\subsection{Hyperparameter Configurations}
\label{app:hyperparameters}
To ensure the reproducibility of our results, we provide the detailed hyperparameter settings used for LIRF training across all experiments. The generative backbone is based on the SiT-B/2 architecture. The DiNO-VAE used for semantic latent embedding operates at a down-sampling factor of 8 with a latent dimension of $d=32$. The refinement interval $\Delta$ is set to 50,000 training steps, which balances computational cost with the necessity of timely manifold densification. We employ a linear decay schedule for the interpolation factor $\lambda$, starting from $\lambda_{\text{start}}=0.8$ at the beginning of training and decreasing to $\lambda_{\text{end}}=0.2$ at the final iteration. The correction magnitude threshold is fixed at $\tau=0.1$.  For the correction operator $\mathcal{C}(\cdot)$, we use $k=3$ nearest neighbors to compute the local reference point $z_{\text{ref}}$. We employ the AdamW optimizer with a constant learning rate of 1e-4 and a total batch size of 32. During training, we monitor the generative performance and report the results from the checkpoint achieving the best FID/Recall.

\subsection{Additional Qualitative Results}
\label{app:add_results}

\paragraph{Impact of the correction strength $\lambda$.}
We visualize the effect of the correction strength $\lambda$ in the operator $\mathcal{C}(\cdot)$ (Definition~\ref{def:correction_operator}) on FFHQ.
We fix a set of generated candidates and decode the corrected latents under different $\lambda$.
As shown in \Cref{fig:lambda_effect}, $\lambda=0$ (no correction) often yields off-manifold candidates that decode into severe structural artifacts and noise.
Increasing $\lambda$ progressively contracts candidates toward the local latent manifold, substantially improving structural coherence while largely preserving the high-level semantic layout. This observation supports the annealed schedule used in the main paper: stronger correction in early training to suppress off-manifold drift, followed by weaker correction to retain diversity as the learned field becomes more reliable.

\paragraph{Latent distribution evolution on MNIST.}
In \Cref{fig:mnist_tsne}, we visualize the evolution of the latent space in the first 10 iterations of the LIRF procedure on the MNIST dataset. Each point represents a sample, where circles correspond to real samples, triangles to generated samples, and squares to corrected samples. As the training progresses, we observe that the generated and corrected samples gradually move towards the real samples, indicating increasing consistency with the target distribution while maintaining diversity in the generated samples.

\paragraph{Qualitative samples on Low-shot benchmarks.}
We provide additional qualitative results on the Low-shot benchmark (Obama, Grumpy Cat, and Panda), each containing only 100 training images.
Following figures show randomly sampled generations from LIRF.
Despite the extreme data scarcity, LIRF produces samples that remain consistent with the target concept while exhibiting substantial variation in pose, expression, viewpoint, and background.
These results visually corroborate the quantitative improvements in Recall reported in Table~\ref{tab:low_shot_results}, indicating better distribution coverage beyond near-duplicates of training examples.

\begin{figure}[h!]
 \centering
\includegraphics[width=0.5\columnwidth]{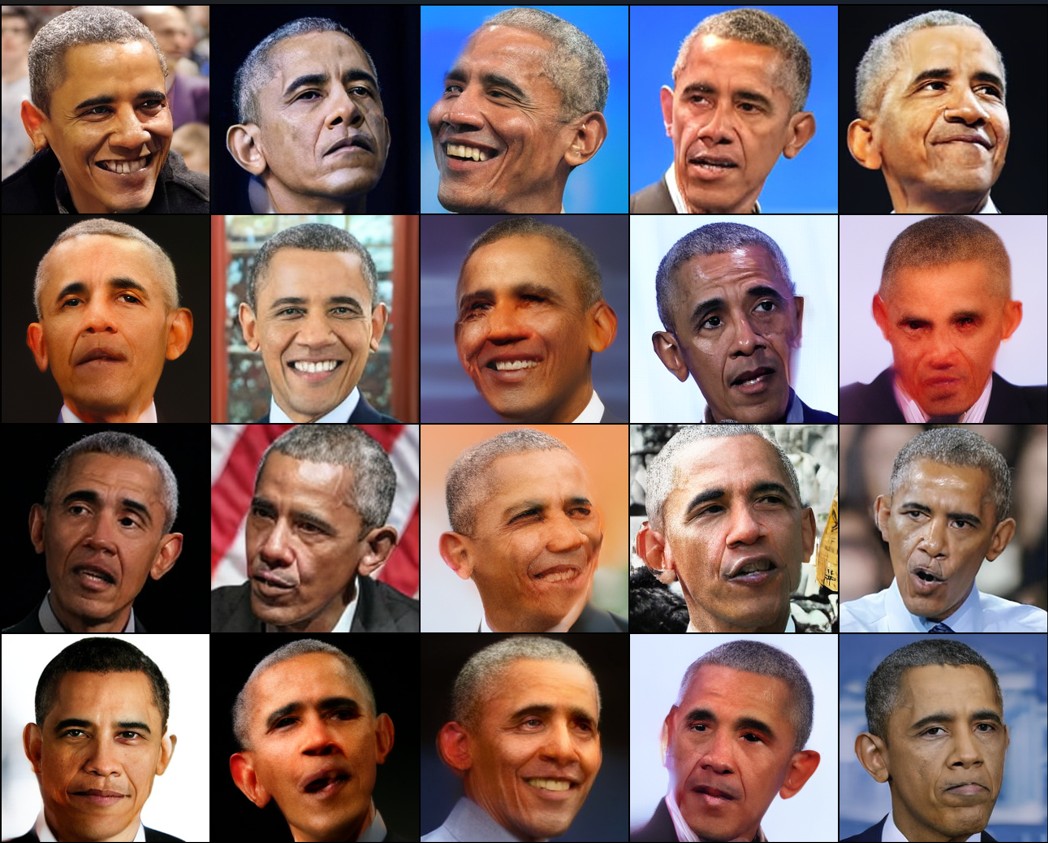}
\caption{Qualitative results on Obama datasets}
     \label{fig:obama}
     \vspace{-0.1in}
 \end{figure}
\begin{figure}[h!]
 \centering
\includegraphics[width=0.5\columnwidth]{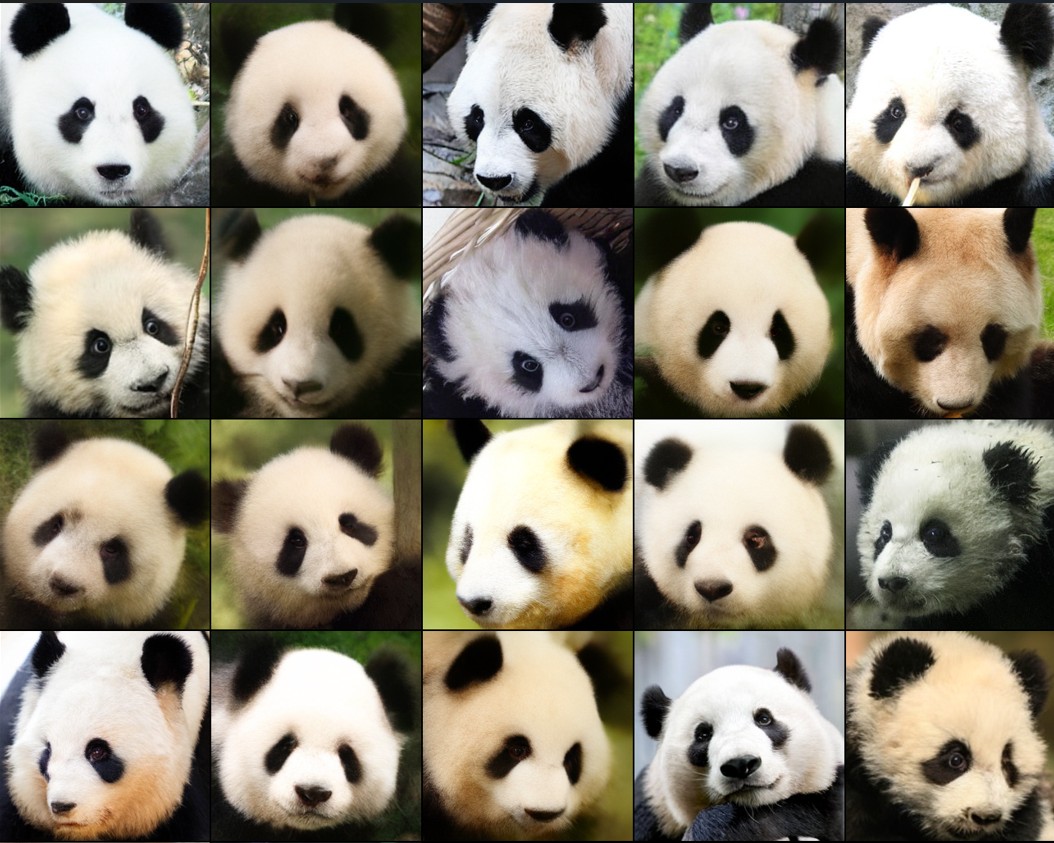}
\caption{Qualitative results on Panda datasets}
     \label{fig:panda}
     \vspace{-0.1in}
 \end{figure} 
\begin{figure}[h!]
\centering
\includegraphics[width=0.5\columnwidth]{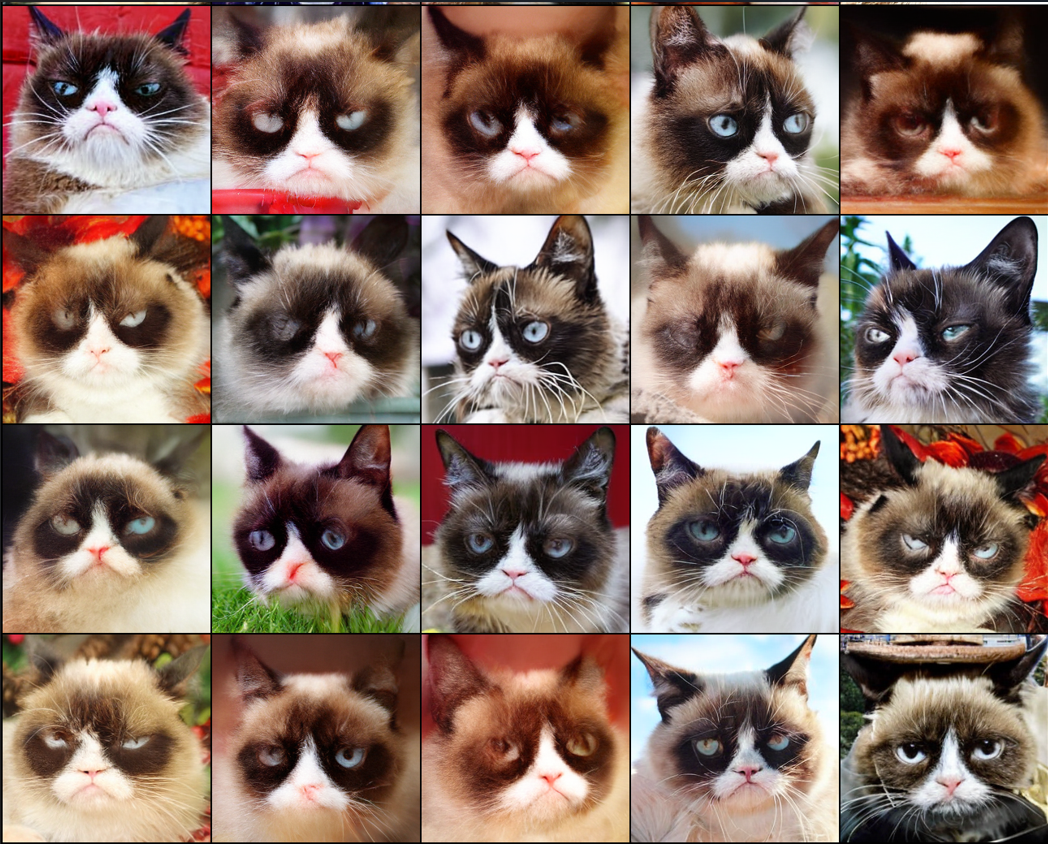}
\caption{Qualitative results on Grumpy Cat datasets}
     \label{fig:cat}
     \vspace{-0.1in}
\end{figure}

\end{document}